\documentclass{article}
\usepackage{PRIMEarxiv}
\usepackage[utf8]{inputenc}

\title{Poly-CAM: High resolution Class Activation Map for Convolutional Neural Networks}
\author{Alexandre Englebert$^{1,2}$, Olivier Cornu$^{2}$ \& Christophe De Vleeschouwer$^{1}$\\
$^{1}$Information and Communication Technologies, Electronics and Applied Mathematics (ICTEAM),\\ UCLouvain, Belgium\\
$^{2}$Service de chirurgie orthopédique et traumatologie, cliniques universitaires Saint-Luc UCL,\\
Neuro Musculo Skeletal Lab (NMSK), Institut de Recherche Expérimentale et Clinique (IREC),\\
université catholique de Louvain, Brussels, Belgium}

\usepackage{natbib}
\usepackage{graphicx}
\usepackage{float}
\usepackage{caption}
\usepackage{subcaption}
\usepackage{amsmath}
\usepackage{amsfonts}
\usepackage{amssymb}
\usepackage{bm}
\usepackage{multirow}
\usepackage{hhline}
\usepackage{hyperref}

\usepackage{algorithm}
\usepackage[noend]{algpseudocode}

\usepackage{comment}

\usepackage[normalem]{ulem}

\def\eqref#1{equation~\ref{#1}}

\def\1{\bm{1}}

\def\vw{{\bm{w}}}

\def\mA{{\bm{A}}}

\def\mM{{\bm{M}}}

\def\mP{{\bm{P}}}

\DeclareMathAlphabet{\mathsfit}{\encodingdefault}{\sfdefault}{m}{sl}
\SetMathAlphabet{\mathsfit}{bold}{\encodingdefault}{\sfdefault}{bx}{n}
\newcommand{\tens}[1]{\bm{\mathsfit{#1}}}
\def\tA{{\tens{A}}}

\def\sN{{\mathbb{N}}}

\newcommand{\R}{\mathbb{R}}

\usepackage{color}
\newcommand{\modcolor}{black}

\begin{document}

\maketitle

\begin{abstract}
The need for Explainable AI is increasing with the development of deep learning. The saliency maps derived from convolutional neural networks generally fail in localizing with accuracy the image features justifying the network prediction. This is because those maps are either low-resolution as for CAM \citep{zhou2016learning}, or smooth as for perturbation-based methods \citep{zeiler2014visualizing}, or do correspond to a large number of widespread peaky spots as for gradient-based approaches \citep{sundararajan2017axiomatic, smilkov2017smoothgrad}. In contrast, our work proposes to combine the information from earlier network layers with the one from later layers to produce a high resolution Class Activation Map that is competitive with the previous art in term of insertion-deletion faithfulness metrics, while outperforming it in term of precision of class-specific features localization.
\end{abstract}

\section{Introduction}

We currently face unprecedented advances in the domain of artificial intelligence, primarily driven by the development of deep neural networks (DNNs). However, in contrast to techniques based on handcrafted features, DNNs often lack transparency and explainability \citep{adadi2018peeking}. 
The need to assess a posteriori the behavior of a model has led to the development of explainable artificial intelligence (XAI) methods, ranging from more transparent models to post-hoc methods (explanation by example of black-box methods), see \citet{samek2021explaining} for a review.

Focusing on convolutional neural networks (CNN), saliency maps visualization has been adopted as a convenient approach to identify the image parts justifying the network prediction. 

Those saliency maps are helpful to check that the predictions of a model are grounded on relevant information. It is indeed known that training convergence alone does not exclude undesired DNN predictions \citep{lapuschkin2019unmasking}, typically because the model has learnt inputs/outputs correlations that do not correspond to the desired meaningful causal relationship. This case is illustrated in Appendix \ref{app:cast_bias}, where a model trained to detect fractures in bone X-rays actually appear to rely on the plaster cast to make its decision rather than on a potential break in the bone.

Alternatively, when sufficiently accurate, the localization of salient features could convince a user that a model works properly, i.e. uses relevant cues, or could even help in identifying the parts of a signal that are relevant to solve a problem, e.g. help a medical doctor in identifying the X-ray visual cues that help to anticipate the evolution of a treatment (Appendix \ref{app:xray_acc}).

Various techniques are available to define class-specific saliency, including perturbations based analysis \citep{zeiler2014visualizing}, gradient based techniques such as Integrated Gradients \citep{sundararajan2017axiomatic} or SmoothGrad \citep{smilkov2017smoothgrad}, and Class Activation Mapping techniques such as Grad-CAM\citep{selvaraju2017grad}, Grad-CAM++\citep{chattopadhay2018grad}, or Score-CAM \citep{wang2020score}.
Class activation maps \citep{zhou2016learning} methods are limited in resolution, while gradient based techniques are generally subject to noise and thus produce saliency maps composed of a large number of widespread peaky spots. \textcolor{\modcolor}{In an attempt to get the best out of both strategies, recent solutions such as Zoom-CAM \citep{shi2021zoom} and Layer-CAM \citep{jiang2021layercam} have proposed to combine the gradients in earlier layers with activations to produce high resolution maps. As shown in our experiments, those maps however inherit some noise from the gradients.}

 Our work introduces a new method to generate high resolution class activation maps, without relying on gradient backpropagation and thus limiting the noisy appearance of the map. In short, this is obtained by multiplexing the high-resolution activation maps available in the early layers of the network with upsampled versions of the class-specific activation maps computed in the last layers of the network. It achieves state of the art performances on faithfulness metrics, and largely improves the localization accuracy of features that explain the network prediction.

\section{Related Work}

Various strategies allow to visualize the image features contributing to the class prediction of a CNN.

Perturbation-based methods use multiple perturbations of the input image, and monitor the changes they induce at the output of the model, to build a saliency map. Those methods range from occlusion techniques, which recursively occlude patches of the input \citep{zeiler2014visualizing}, to more elaborated perturbations such as Randomized Input Sampling for Explanation of Black-box Models (RISE) \citep{petsiuk2018rise}, which generate many random masks normalized between 0 and 1, to be weighted by the class-specific softmax output obtained when the input image is multiplied by the mask. 
These methods result in a smooth mask and are computationally intensive, especially when the resolution of the saliency map increases.

Gradient based methods use the back-propagated gradient of the neural network to identify regions in the input image that largely impact the prediction. Those methods however suffer from gradient shattering \citep{bahdanau2014neural}, which results in a noisy saliency map. To mitigate this problem, a variety of approaches have been implemented to smooth out the gradient signal. They include Integrated Gradient \citep{sundararajan2017axiomatic}, which integrates the gradient for multiple interpolations between a baseline and the input image, or SmoothGrad \citep{smilkov2017smoothgrad}, which averages gradient for multiple perturbed variations of the input image.

Class Activation Map \citep{zhou2016learning} was initially designed as a linear combination of activations from the last convolutional layer, weighted by the parameters of a fully connected classifier, taking as input the global average pooling of each channel in this last convolutional layer. Multiple methods were introduced based on alternative definitions of the weights. Grad-CAM \citep{selvaraju2017grad} and Grad-CAM++ \citep{chattopadhay2018grad} define the weights based on gradients backpropagated upto the last convolutional layer. Score-CAM defines the weight of a channel 
based on the softmax output associated to the target class when probing the model with a version of the image masked by the activation channel. In a sense, it mixes CAM methods with perturbations methods \citep{wang2020score}. Overall, Class Activation Map methods are less noisy than gradient-based methods but are coarser, with a resolution limited to the resolution of the last convolution layer.

\textcolor{\modcolor}{To produce CAM at higher-resolution,}
\citet{tagaris2019high} propose to \textcolor{\modcolor}{train an expansion network}, but this has the drawback to require a specific training for each model to analyze 
\citep{ronneberger2015u}. At the time of writing, this expansion network is only defined for DenseNet \citep{huang2017densely} and trained on a subset of animal labels of ImageNet \citep{ILSVRC15}, available at https://github.com/djib2011/high-res-mapping. \textcolor{\modcolor}{Very recently, Layer-CAM \citep{jiang2021layercam} and Zoom-CAM \citep{shi2021zoom} have proposed to generate high-resolution CAM by combining back-propagated gradients with activations from multiple layers, using element-wise multiplications. 
The two methods improve the resolution of the maps but also inherit some noise from the gradients.}

Our paper proposes an original method to compute high-resolution activation maps without using gradients, nor requiring to train a specialized network. Our primary contribution is a process to leverage the activation from multiple layers to increase the resolution of the Class Activation Map up to the one of the input image. Our second contribution is a new way to compute the weights of the activation maps, 
It introduces a dual strategy compared to the approach introduced by Score-CAM \citep{wang2020score}, and proposes to get the best out of both strategies by merging them.

\section{Proposed Poly-CAM Approach}

This section presents the core contribution of our work. Section \ref{notation} introduces the notations and variables required in the rest of the text, while Section \ref{previous_work_formulas} reviews the formal definition of the conventional Class Activation Map method, which serves as a baseline to our work. Section \ref{pcam_def} then introduces our Poly-CAM approach, which proposes to generate a high resolution class activation map by recursively multiplexing the high-resolution activation maps available in the early layers of the network with upsampled versions of the class-specific activation maps computed in the last layers of the network. Eventually, Section \ref{weights_definition} introduces three different methods to associate a weight to each layer activation channel. All methods measure how the output of the network is affected when masking/unveiling the input based on the channel activation.

\subsection{Notations}\label{notation}

Let $f_\Theta(X)$ denote the prediction of a CNN with parameters $\Theta$ when the image $X$ is provided as input. In the following, for conciseness and because we are interested in analyzing a trained network (parameters $\Theta$ are fixed), we omit $\Theta$, and just use $f(X)$ to refer to the CNN prediction associated to $X$. In the following, $f(X)$ is a vector, defined by the output of a softmax.
$f_c(X)$ denotes the component of $f(X)$ corresponding to the class c.\\
$\tA_l$ denotes the activation tensor of the $l^{th}$ convoluional layer, $1 \leq l \leq L$, while $\mA^k_l$ refers to the activations of the k-th channel of layer $l$.\\
$s_l$ denotes the subsampling factor of layer $l$ compared to the input. It corresponds to the product of stride and pooling factors encountered between the input and layer $l$.\\
$\uparrow_{bi}(\mM, s)$ defines a bilinear upsampling of a matrix $\mM \in \R^{m \times n}$ by a factor $s \in \sN$.\\
$\downarrow_{av}(\mM, s)$ denotes a 2D average pooling on any matrix $\mM \in \R^{m \times n}$ with a stride $s \in \sN$.\\
$u(\mM)$ linearly maps the value range of the elements in matrix $\mM$ to the unit interval. \\
$\oslash$ denotes the element-wise division operator, while $\odot$ denotes the element-wise product operator.\\
$LNorm(\mM, s)$ is a local normalisation operator. It partitions the matrix $\mM \in \R^{m \times n}$ in a set of non overlapping blocks of size $s \times s$, with $s \in \sN$, and divides each matrix element by the mean value of its corresponding block. Formally, using the above notations,
        \begin{equation}\label{eq:lnorms} 
            LNorm(\mM, s) = \mM \oslash \left(\uparrow_{bi}(\downarrow_{av}(\mM, s), s)\right).
        \end{equation}
$ReLU$ denotes the rectified linear units \citep{dahl2013improving}.

\subsection{Activation Maps in previous work}\label{previous_work_formulas}
In \citep{zhou2016learning} $CAM^c_{l}$, the Class Activation Map associated to a target class $c$ and a layer $l$ is defined as
\begin{equation}\label{eq:cam}
    CAM^c_{l} = ReLU(\sum_{k} w_{l,k}(c) \mA^k_l ),
\end{equation}
\textcolor{\modcolor}{with $\mA^k_l$ denoting the $k^{th}$ activation map of the $l^{th}$ convolutional layer, $l \in 1,...,L$, and $w_{l,k}(c)$ being a scalar weighting factor. 
}
Most of the CAM-based methods \citep{selvaraju2017grad, chattopadhay2018grad, smilkov2017smoothgrad, wang2020score, wang2020ss, naidu2020cam} adopt this formula. They differ in the way they define the weighting factors, and generally only consider it for the last convolutional layer ($l=L$). 
\textcolor{\modcolor}{Alternatively, Zoom-CAM and Layer-CAM have proposed to combine activation maps from multiple layers, using gradients as dense weighting factors.} Our work also combines multiple activation maps, \textcolor{\modcolor}{but does it without back-propagated gradients, thereby managing to produce high-resolution saliency maps without inheriting the noise from the gradient}.

\subsection{Our proposed Poly-CAM}\label{pcam_def}
\begin{figure}
	\centering
	\includegraphics[width=0.8\textwidth]{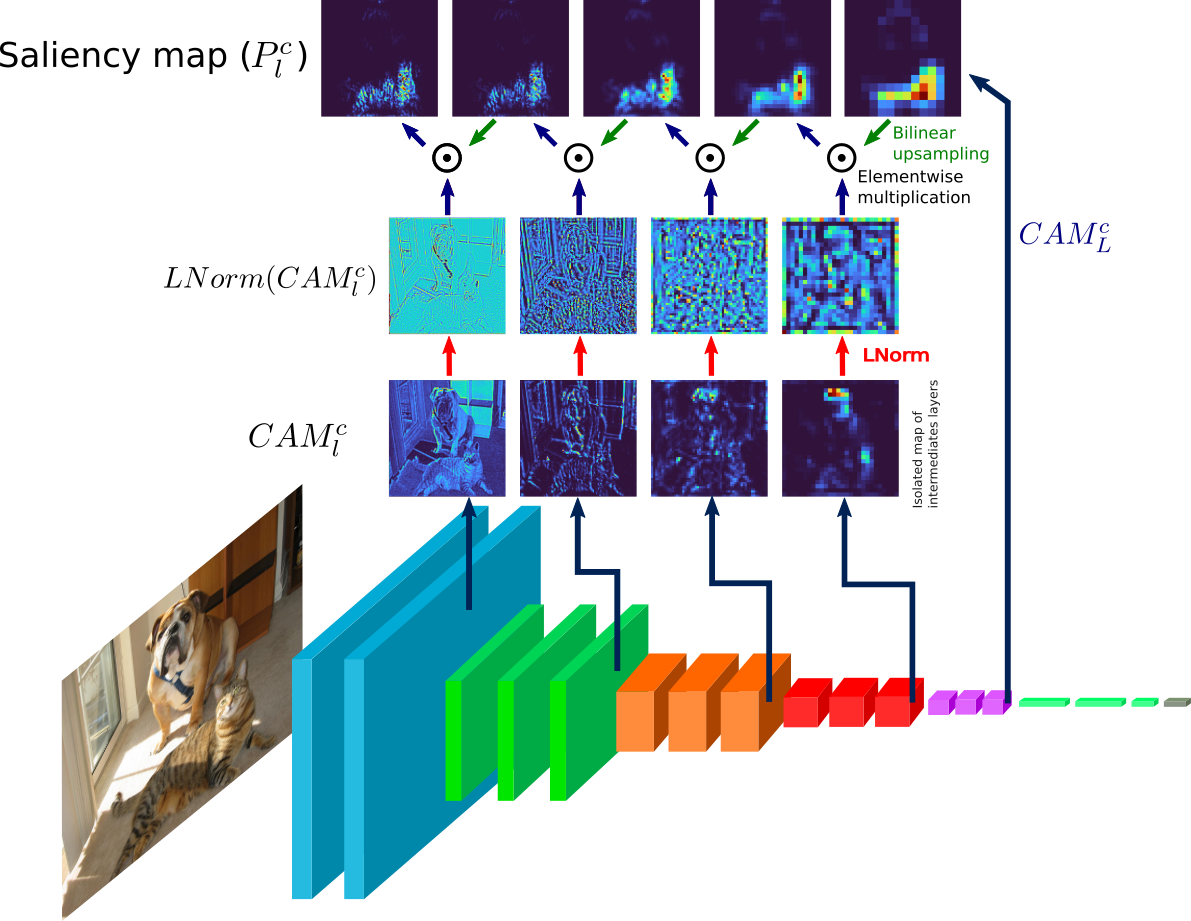}
	\caption{Our Poly-CAM process: the upsampled version of the saliency map in layer $l$ is  tuned based on the class activation map of layer $l-1$. Image samples correspond to the 'cat' class, and are computed from VGG16.}
	\label{fig:polycam_process}
\end{figure}

Our method leverages information from multiple layers to produce a high resolution Class Activation Map. Similar to other CAM-based techniques, it builds on the linear combination of activation maps, but combines them through a backward recursive procedure, \textcolor{\modcolor}{ as depicted in Figure \ref{fig:polycam_process}.\\}
\textcolor{\modcolor}{Letting $P^c_{l}$ denote the class-specific saliency map associated to class $c$ in the $l^{th}$ layer, the recursive process works as follows. In the initial step, the saliency map $P^c_{L}$ is defined to be equal to the conventional $CAM^c_L$ saliency map, as derived from equation (\ref{eq:cam}). Then, at each recursive step, an upsampled version of $P^c_{l+1}$ is tuned (or modulated) by a locally normalized version of the activation map in the $l^{th}$ layer. Mathematically, we have:
}



\begin{equation}
\color{\modcolor}
P^c_l = 
\left \{
    \renewcommand\arraystretch{2}
    \begin{array}{ll}
        ReLU(\sum_{k} w_{l,k}(c) \mA^k_l ) & \mbox{for } l = L \\
        LNorm\left(ReLU\left(\sum_{k} w_{l, k}(c) \mA^{k}_{l}  \right), \frac{s_{l+1}}{s_l}\right) \odot \uparrow_{bi}\left(\mP^c_{l+1}, \frac{s_{l+1}}{s_l}\right) & \mbox{for } 1 \leq l \leq L-1 \\
    \end{array}
\right .
\label{eq:poly-cam}
\end{equation}

with $\text{w}_{l, k}(c)$ denoting the \textcolor{\modcolor}{class-specific weighting factor associated to $A^k_l$,
as defined in Section \ref{weights_definition}.}\\
\textcolor{\modcolor}{
Intuitively, Equation (\ref{eq:poly-cam}) can be understood based on the following two observations.
First, the element-wise multiplication, between the upsampled (and thus smooth) saliency map of layer $l+1$ and the activation map in layer $l$, aims at restricting the large saliency values in layer $l+1$ to the locations that are activated in layer $l$.
Second, the local normalization ($LNorm$) of the activation map aims at preserving the spatial distribution of saliency across the layers. It ensures that an image block with large saliency in layer l+1 has also a large saliency in layer l, even if the level of activation in this block is small compared to the rest of the image. This is meaningful since the backpropagated saliency should be predominant to assign a saliency level to a spatial region in layer l, while the activation in layer l should simply control the increase in resolution, by tuning the smooth saliency signal inherited from coarser layers based on the local variations of the activation map.  
Our ablation study in Appendix \ref{app:lnorm_ablation} confirms the critical role played by the LNorm operator.
}

\subsection{Activation map weight definition}\label{weights_definition}

This section presents different alternatives to define the weights $w_{l,k}(c)$, used in \eqref{eq:poly-cam}. 

\vspace{-0.2cm}

\paragraph{Channel-wise Increase of Confidence}
 Score-CAM \citep{wang2020score}, SS-CAM \citep{wang2020ss} and IS-CAM \citep{naidu2020cam} define $w_{l,k}(c)$ based on the so-called Channel-wise Increase of Confidence (CIC), which estimates how the spatial support of the activation map $A^k_l$ contributes to the softmax output $f$. 
 Formally, the channel-wise increase is denoted $\vw_{l, k}^+$, and is defined as:
\begin{equation}\label{eq:cic}
    \vw_{l, k}^+ = f\left(X \odot u\left(\uparrow_{bi}\left(\mA^k_l, s_l\right)\right)\right) - f\left(X_b\right),
\end{equation}
with $\odot$ denoting the pixel-wise product, and $X_b$ referring to a baseline image. Previous works have considered baselines that are uniform black, uniform grey, or a blur version of $X$. In the following, $f_c(X_b)$ is set to zero in all experiments. 

Our work proposes two extensions of \eqref{eq:cic}, respectively to measure how the softmax output decreases when masking a fraction of the input, and to sum-up the increase and the decrease associated to the unveiling and the masking of the input. Those new weights are defined as follows.

\paragraph{Channel-wise Decrease of Confidence}

The Channel-wise Decrease of Confidence ($CDC$) is a dual notion compared to $CIC$. Instead of measuring the increase of softmax output when the part of the input corresponding to non-zero $\mA^k_l$ is unveiled and the remaining is masked, CDC measures the decrease of softmax output when the part of the input corresponding to $\mA_k^l$ is masked. The intuition is that an important part of the input for any class $c$ not only increase the output when shown, but should also decrease it when hidden. Formally,
\begin{equation}\label{eq:cdc}
    \vw_{l, k}^- = ReLU\Big(f(X) - f\left(X \odot \left(1 - u\left(\uparrow_{bi}\left(\mA^k_l, s_l\right)\right)\right)\right)\Big)
\end{equation}
$ReLU$ is applied to only keep the activation maps that decreases the output when removed.

\paragraph{Channel-wise Variation of Confidence}

By combining the $CIC$ with the $CDC$, the Channel-wise Variation of Confidence ($CVC$) is defined. Formally,
\begin{equation}\label{eq:cvc}
    \vw_{l, k}^{\pm} = ReLU\Big(f\left(X\right) + f\left(X \odot u\left(\uparrow_{bi}\left(\mA^k_l, s_l\right)\right)\right) - f\left(X \odot \left(1 - u\left(\uparrow_{bi}\left(\mA^k_l, s_l\right)\right)\right)\right)\Big)
\end{equation}
The Channel-wise Variation of Confidence is either influenced by the ability of the activation map to increase the softmax output when inserted but also decrease it when removed.

\section{Experiments}\label{material_and_methods}

Three variants of the Poly-CAM method introduced in Section \ref{pcam_def} are considered, depending on whether $\vw_{l, k}$ is defined to be equal to $\vw_{l, k}^{+}$ (PCAM$^+$), $\vw_{l, k}^{-}$ (PCAM$^-$) or $\vw_{l, k}^{\pm}$ (PCAM$^\pm$).

We follow the assessment method described in \citep{petsiuk2018rise} and \citet{chattopadhay2018grad} to evaluate our proposal.
Datasets, networks, and baseline methods are presented in Section~\ref{dataset_methods}.
Quantitative assessment of the saliency maps is considered in Section~\ref{faithfulness_assessment}, while a qualitative and visual assessment is presented in Section~\ref{visual_assessment}.

\subsection{Experimental set-up and saliency map baselines}\label{dataset_methods}

For theses evaluations, 2000 images were randomly selected from the 2012 ILSVRC validation set \citep{ILSVRC15}.
The images are scaled to 224x224x3 pixels and normalized to the same mean and standard deviation as the ImageNet \citep{ILSVRC15} training set (mean vector : [0.485, 0.456, 0.406], standard deviation vector [0.229, 0.224, 0.225]).
The models used for faithfulness evaluation are VGG16 \citep{simonyan2014very} and ResNet50 \citep{he2016deep}, both pretrained from the PyTorch model zoo. The analysis considers, as reference baselines, Grad-CAM \citep{selvaraju2017grad}, Grad-CAM++ \citep{chattopadhay2018grad}, Smooth Grad-CAM++ \citep{omeiza2019smooth}, Score-CAM \citep{wang2020score}, SS-CAM \citep{wang2020ss}, IS-CAM \citep{naidu2020cam}, Zoom-CAM \citep{shi2021zoom}, Layer-CAM \citep{jiang2021layercam}, Occlusion \citep{zeiler2014visualizing}, Input X Gradient \citep{shrikumar2016not}, Integrated Gradient \citep{sundararajan2017axiomatic}, SmoothGrad \citep{smilkov2017smoothgrad} and RISE \citep{petsiuk2018rise}. The implementations for these methods are the ones from Captum \citep{kokhlikyan2020captum} for Integrated Gradient, SmoothGrad and Occlusion, from \url{https://github.com/eclique/RISE} for RISE,  \textcolor{\modcolor}{from \url{https://github.com/X-Shi/Zoom-CAM} for Zoom-CAM} and from torchcam \citep{torcham2020} for all the other CAM-based methods.


For SS-CAM, IS-CAM, LayerCAM and ZoomCAM, SmoothGrad and IntegratedGradien, the parameters recommended by the authors or set as default in the reference implementation have been used when available.
\footnote{It means 35 input perturbations (with a $\sigma=2$ Gaussian noise) for SS-CAM, 50 input perturbations (with a $\sigma=1$ Gaussian noise) for SmoothGrad, 10 interpolation steps for IS-CAM, and 50 for IntegratedGradient.
For Layer-CAM, the layers corresponding to a change in resolution were used, and recommended scaling has been applied to the first two layers. 
For Zoom-CAM, all the layers/blocks were fused for VGG16 and ResNet50.}
For Occlusion, the size of occlusion patch was set to (64, 64) with a stride of (8, 8) as used by \citep{petsiuk2018rise}. For RISE, 6000 masks were used.


For the Poly-CAM methods (PCAM$^+$, PCAM$^-$, PCAM$^\pm$), the layers corresponding to a change in resolution were used. It corresponds to [block1\_conv2,\ block2\_conv2,\ block3\_conv3,\ block4\_conv3,\ block5\_conv3] for VGG16, and [conv1\_1,\ conv2\_3,\ conv3\_4,\ conv4\_6,\ conv5\_3] for ResNet50.

\subsection{Faithfulness Assessment}\label{faithfulness_assessment}

There is still a lack of consensus regarding the metrics to assess the relevance of saliency maps for explainability \citep{poursabzi2021manipulating}. The ability to segment the semantic object justifying the class label was used as a proxy to evaluate saliency maps \citep{selvaraju2017grad}, but it has been observed in \citet{petsiuk2018rise} that the segmentation mask might not be correlated to the discriminant visual features justifying the class label. The same authors also introduced the insertion and deletion metrics to measure the increase (resp. decrease) in the class softmax output when inserting (resp. removing) pixels in decreasing order of saliency. 
Theses metrics are largely used in recent works \citep{wang2020score, wang2020ss, naidu2020cam}. They are thus considered in this work, \textcolor{\modcolor}{even if they remain arguable, as discussed in the following sections.}

\subsubsection{Metrics definition}

\paragraph{Insertion}
Formalised by \citet{petsiuk2018rise}, the insertion metric measures how fast the model softmax output increases when adding the salient image pixels to a baseline image. The pixels are of a baseline (uniform black/grey or highly blurred version of the image) are replaced by the image ones in decreasing order of saliency map value, allowing to define the faithfulness metric as the area under the curve of the class softmax output, observed as a function of the proportion of pixels inserted. The higher the metric the better. 
A blurred version of the image is generally preferred to a uniform color baseline, since it prevents the introduction of sharp artificial edges that might disrupt the predictions.

\paragraph{Deletion}
The deletion metric \citep{petsiuk2018rise} measures the decrease in class softmax output while removing pixels. The pixels are progressively removed from the image and replaced by a baseline using the same logic as the insertion metric.
The intuition behind the deletion metric is that removing the more important pixels in the saliency map should more rapidly decrease the softmax output of the target class prediction. The lower the metric the better.

\paragraph{Insertion - Deletion}
A good explanation map should have a good insertion map and a good deletion map. A combination of the two scores is thus proposed here by computing the difference between insertion (higher is better) and deletion (lower is better) to obtain a combined score that should be maximized. 

\textcolor{\modcolor}{It is worth noting that, despite they are  widely used, those metrics have some drawbacks. In particular, the insertion and deletion procedures are likely to result in outliers, i.e. in images that do not match the distribution of natural images. To mitigate this problem, as most previous works, we have used blurred baselines when implementing the insertion/deletion process. However, it is not sufficient to ensure that a high correlation between the network prediction and the salient pixels means that those pixels exactly correspond to the whole set of class-discriminant visual features.
Hence, those quantitative metrics should never be used as a substitute for a visual assessment of the saliency map, which remains the golden standard when comparing saliency map generation methods.} 

For the above metrics, 224 steps were performed with 224 pixels inserted or removed at each step. The used baseline for all the metrics is a blurred version of the input image using a Gaussian kernel of size 11x11 with a sigma of 5. 

\subsubsection{Quantitative assessment}

\begin{table}[ht]
\caption{Faithfulness metrics for CAM-based methods \textcolor{\modcolor}{(see Table \ref{table_comparison_other} for full version with all methods)}}
\centering
\begin{tabular}{l|ccc|ccc}
\hline
\multirow{2}{*}{Method} & \multicolumn{3}{c}{VGG16}& \multicolumn{3}{c}{ResNet50}\\
\hhline{~------}
 &  Insertion     &   Deletion     &     Ins-Del &  Insertion     &   Deletion     &     Ins-Del \\ \hline
GradCAM & 0.58 & 0.18 & 0.40 & 0.65 & 0.31 & 0.35\\
GradCAM++ & 0.57 & 0.19 & 0.38 & 0.65 & 0.31 & 0.34\\
SmoothGradCAM++ & 0.54 & 0.21 & 0.33 & 0.63 & 0.32 & 0.30\\
ScoreCAM & 0.59 & 0.19 & 0.40 & 0.65 & 0.31 & 0.34\\
SSCAM & 0.50 & 0.23 & 0.27 & 0.59 & 0.36 & 0.24\\
ISCAM & 0.59 & 0.19 & 0.40 & 0.65 & 0.32 & 0.33\\
ZoomCAM & 0.60 & \textbf{0.14} & \textbf{0.46} & 0.66 & 0.29 & 0.37 \\
LayerCAM & 0.58 & 0.14 & 0.44 & 0.65 & 0.30 & 0.35\\\hline
PCAM$^+$ (ours) & 0.58 & 0.17 & 0.41 & \textbf{0.67} & 0.29 & 0.38\\
PCAM$^-$ (ours) & 0.60 & 0.16 & 0.45 & 0.66 & \textbf{0.27} & \textbf{0.39}\\
PCAM$^{\pm}$ (ours) & \textbf{0.61} & 0.15 & \textbf{0.46} & \textbf{0.67} & 0.28 & \textbf{0.39}\\
\hline
\end{tabular}
\label{table_comparison_cam}
\caption*{Insertion (higher is better), deletion (lower is better) and insertion-deletion (higher is better) with VGG16 and ResNet50 on the 2012 ILSVRC validation set. Comparison of our Poly-CAM methods with Grad-CAM \citep{selvaraju2017grad}, Grad-CAM++ \citep{chattopadhay2018grad}, Smooth Grad-CAM++ \citep{omeiza2019smooth}, Score-CAM \citep{wang2020score}, SS-CAM \citep{wang2020ss}, IS-CAM \citep{naidu2020cam}, \textcolor{\modcolor}{Zoom-CAM \citep{shi2021zoom}, Layer-CAM \citep{jiang2021layercam}. 
}}
\end{table}

\vspace{-0.2cm}

Table \ref{table_comparison_cam} compares the faithfulness metrics for all CAM-based methods.  The results related to gradient and pertubation methods are presented in Appendix, Table \ref{table_comparison_other}. Plots of the softmax output as a function of the amount of inserted pixels are presented in Appendix \ref{app:curves}.\\
Among the three Poly-CAM variants, PCAM$^\pm$ gives the best results compared to PCAM$^+$ and PCAM$^-$ for all metrics on VGG16. On ResNet50, PCAM$^\pm$ gives a insertion metric similar to PCAM$^+$ and superior to PCAM$^-$, while PCAM$^-$ gives a better deletion metric compared to PCAM$^+$ and PCAM$^\pm$. For insertion-deletion on ResNet50, PCAM$^\pm$ and PCAM$^-$ are on par.\\
Interestingly, the insertion metrics of PCAM$^\pm$ is systematically better than  all other CAM-based approaches.
Compared to the non-CAM methods, the PCAM$^\pm$ method gives similar or better insertion results than perturbation or gradient methods, respectively. 
\\
\textcolor{\modcolor}{In terms of deletion, PCAM$^\pm$ tends to perform better than most other CAM-based methods, but appears to be weaker than gradient-based methods.} InputXGrad and IntegratedGradient achieve at the same time very poor results on the insertion metric and thus have a poor insertion-deletion. This is not surprising since gradient-based methods give lots of importance to the parts of the input that largely impact the loss and thus the output. As a consequence, the deletion metric is (trivially) good for those methods since this metric measures the decrease of output when important parts are removed from the input. The poor insertion metric however reveal that the parts that are considered as being important by gradient methods are not sufficient to explain the network prediction. \textcolor{\modcolor}{ This observation reveals the limits of the metrics when applied to dissimilar kinds of techniques.} 

\subsection{Visual assessment}\label{visual_assessment}

This section assesses our method visually. 
Saliency maps were generated for all the baseline methods (see Section \ref{dataset_methods}) on the 2000 selected images using VGG16 model. For the Poly-CAM methods, saliency maps where also generated for each target layer. An interactive interface is provided as a jupyter notebook in supplementary material, in addition to the source code, to allow an easy visualisation of any of the saliency maps generated in our experiments.
Section \ref{polycam_vs_polycam} compares the three Poly-CAM variants, as a function of the layer index and targeted class.
A comparison with previous works is shown in Section \ref{polycam_vs_other}.
In Section \ref{failure_cases}, PCAM$^\pm$ is considered to explain VGG16 misclassifications. 

\subsubsection{PCAM variants}\label{polycam_vs_polycam}

\textcolor{\modcolor}{PCAM produces saliency maps at various resolutions. Figure \ref{fig:polycam_process} shows how Poly-CAM progressively refines the last layer saliency map through a backward recursive strategy. More examples of intermediate maps obtained along this recursive process are presented in Appendix \ref{app:pcam_isolated_cam}.} We observe that the structures are coarse at block5\_conv3, to gain in accuracy when accounting for earlier network layers, doubling the resolution at each step. The elements highlighted by the three variants are similar on the majority of images. However, variations appear on some images, PCAM$^-$ highlighting more frequently contextual elements compared to PCAM$^+$ (and PCAM$^\pm$ sitting between the two). Intuitively, this can be understood by the fact that the appearance of those contextual features in a baseline image does not help in classifying the image correctly, while their removal from the original image penalizes the classification. 

\begin{figure}
	\centering
	\includegraphics[width=0.9\textwidth]{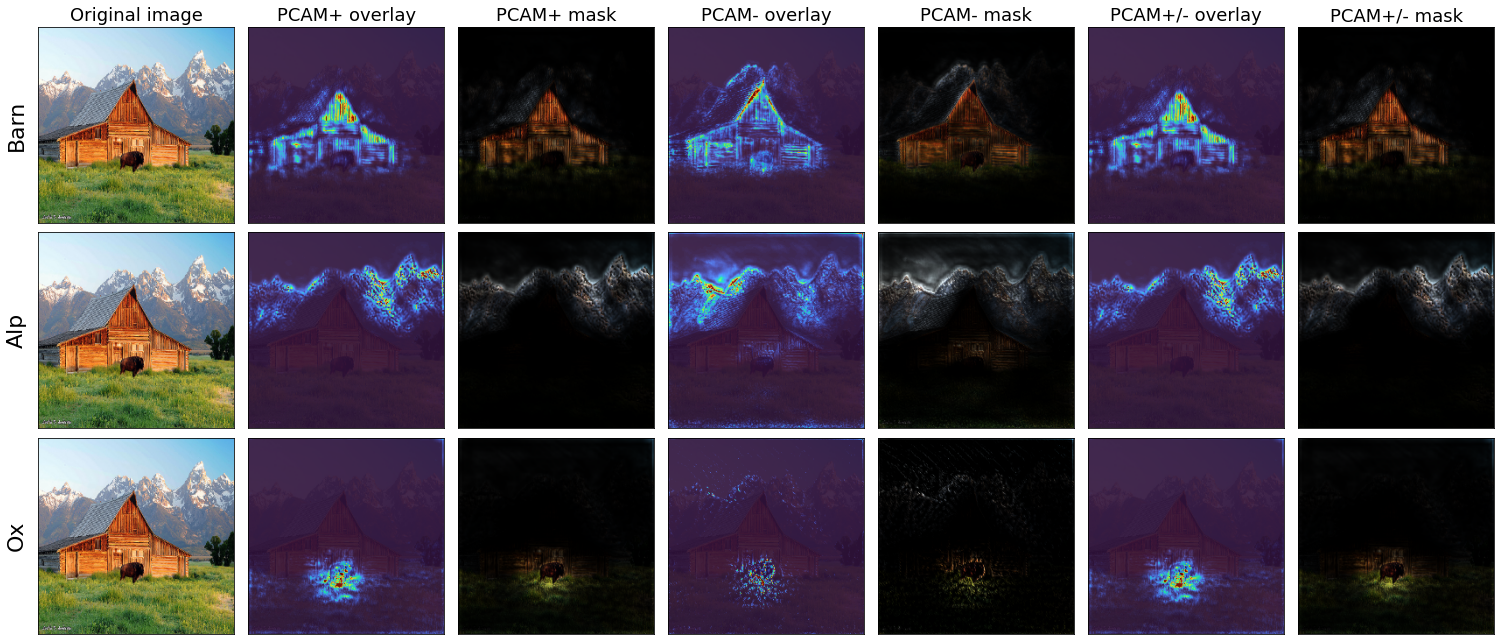}
	\caption{Multiple classes on Poly-CAM. The different classes as correctly determined by the three variants. We can note that PCAM$^-$ is less specific than the other methods (part of the mountain is attributed to the barn), while PCAM$^+$ is the most specific.}
	\label{fig:multiclasses}
\end{figure}

All Poly-CAM variants are class specific as displayed in Figure \ref{fig:multiclasses}, where the saliency maps associated to the the Barn, Alps and the Ox classes are clearly distinct,  with a level of accuracy close to segmentation. It is worth noting that PCAM$^+$ is more specific in highlighting the part of the image related to the class of interest. This is in line with the above observation that PCAM$^-$, and a bit less PCAM$^\pm$, are stronger in highlighting contextual information (see more examples in Appendix \ref{app:class_spe}).

\subsubsection{Comparison with previous works}\label{polycam_vs_other}

\begin{figure}
	\centering
	\includegraphics[width=\textwidth]{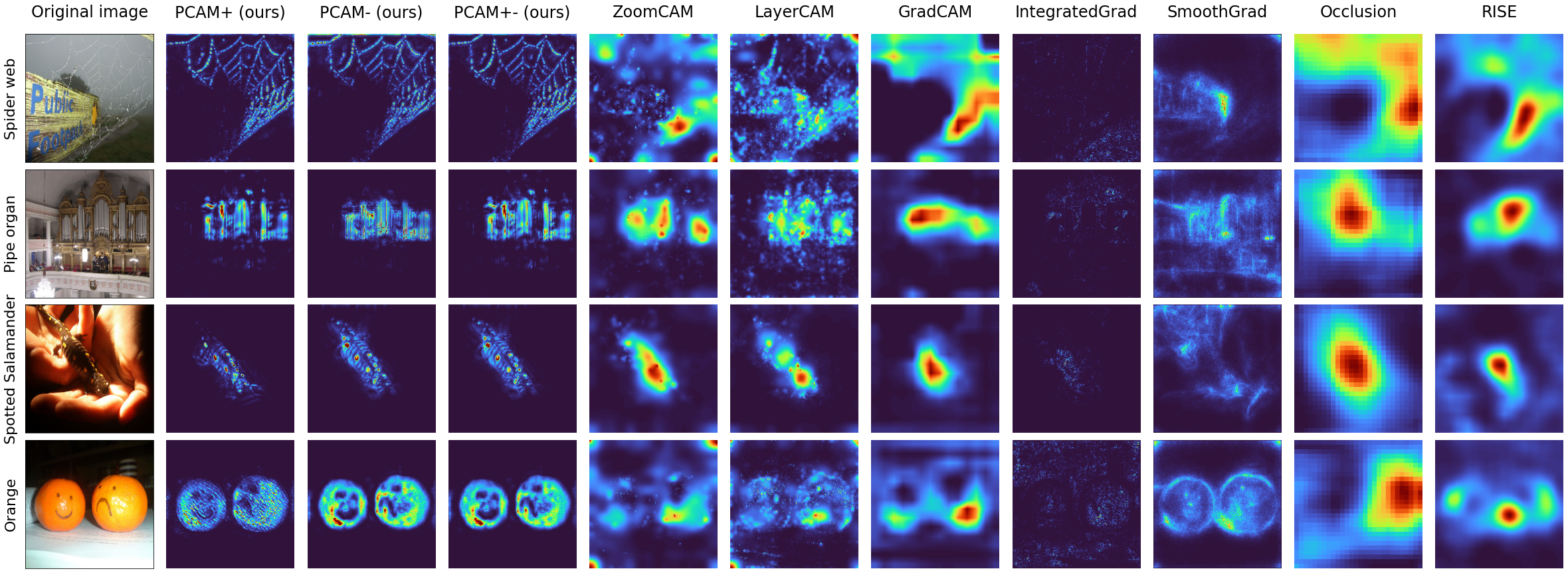}
	\caption{Visual comparison of methods \textcolor{\modcolor}{(see Figure \ref{fig:methods_comparison_full} in appendix for full version)}. The compared methods are our three variants of Poly-CAM (PCAM$^+$, PCAM$^-$, PCAM$^\pm$), \textcolor{\modcolor}{Zoom-CAM \citep{shi2021zoom}, Layer-CAM\citep{jiang2021layercam},} Grad-CAM \citep{selvaraju2017grad}, IntegratedGradient \citep{sundararajan2017axiomatic}, SmoothGrad \citep{smilkov2017smoothgrad}, Occlusion \citep{zeiler2014visualizing} and RISE \citep{petsiuk2018rise}. A description of this figure is available in Section \ref{polycam_vs_other}}
	\label{fig:methods_comparison}
\end{figure}

Saliency maps have been produced for all baseline methods \textcolor{\modcolor}{listed in Section \ref{dataset_methods}}.
A sample of this comparison is presented in Figure \ref{fig:methods_comparison}. For ease of view, this figure restricts to Grad-CAM, Score-CAM, Integrated Gradient, SmoothGrad, Occlusion, RISE and the three Poly-CAM variants. A comparison with all the methods listed in Section \ref{material_and_methods} can be found in Appendix \ref{app:methods_comparison_full}, Figure \ref{fig:methods_comparison_full}.
We can see that Poly-CAM methods accurately identify quite relevant elements in the image like a spider net or the pipes of an organ.  CAM and perturbation methods cannot achieve this level of precision, gradient base methods highlight elements of the image that are not related to the class \textcolor{\modcolor}{, while Zoom-CAM and Layer-CAM increase the resolution but are more noisy, halfway between gradient and more classical CAM-based methods}. The spotted salamander is highlighted by all methods but the Poly-CAM methods are the only ones to identify the spots. The oranges are identified by Poly-CAM methods while the smile sketched on them is correctly excluded. This is in contrast with other methods that are either too low resolution, or do not exclude the smile, while SmoothGrad seems to give more importance to the smile than to the texture of the orange.

\subsubsection{Classification failure explanation}\label{failure_cases}

\begin{figure}
	\centering
	\begin{subfigure}{\textwidth}
	    \centering
		\includegraphics[width=0.7\textwidth]{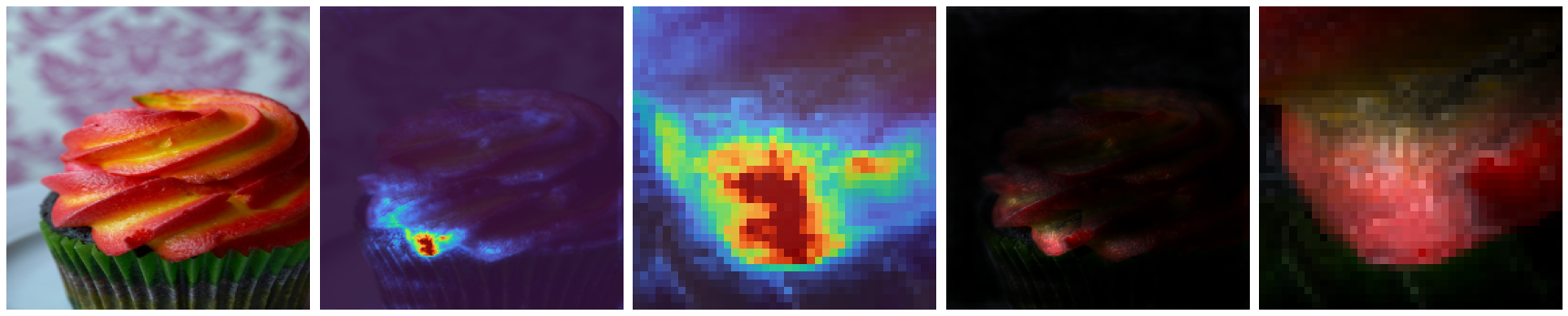}
		\caption{Bakery misclassified as strawberry}
	\end{subfigure}\label{fountain}
	\begin{subfigure}{\textwidth}
	    \centering
		\includegraphics[width=0.7\textwidth]{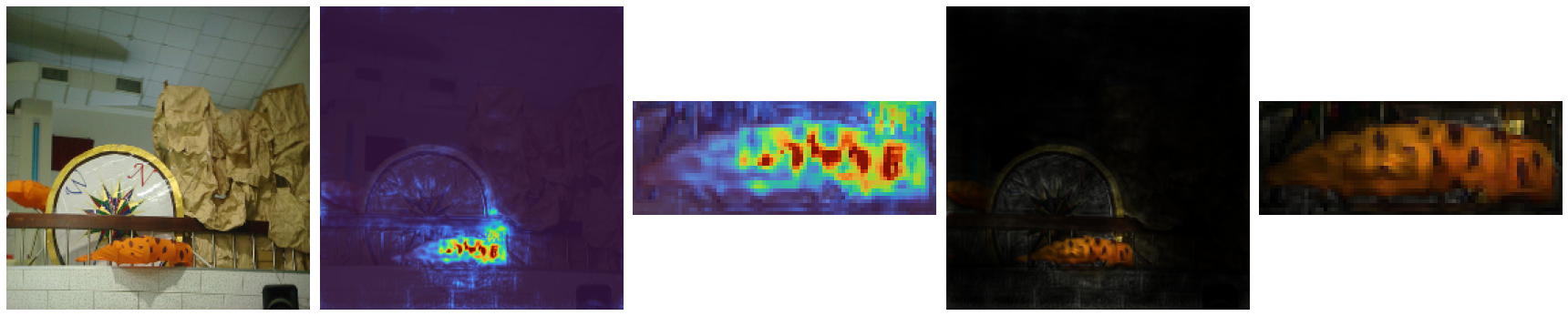}
		\caption{Compass classified as chainsaw}
	\end{subfigure}
    \caption{\textbf{Misclassification explanations.} From right to left: original image, attention map, masked image, and their zoomed versions. (a) Image is misclassified as strawberry. The saliency map explains the wrong classification by highlighting a visual stimuli looking like a strawberry. (b) Image is missclassified as chainsaw, as explained by the saliency map.}
	\label{fig:misclassifications}
\end{figure}

PCAM$^\pm$ is used to produce explanation maps for images misclassified by VGG16, to show the ability of Poly-CAM to explain the reasons why the model made a mistake. Images with misclassification were selected in the dataset. The image samples presented in Figure \ref{fig:misclassifications} have been chosen so that the misclassification is not related to a similar class (e.g. a golden retriever classified as a Labrador retriever) or to another object present in the image. As illustrated in Figure \ref{fig:misclassifications} looking at the image masked by the PCAM$^\pm$ saliency map helps in understanding the misclassification. 
In Figure \ref{fig:misclassifications}(a), we can identify the strawberry seen by the model, when using the saliency map as a mask. In Figure \ref{fig:misclassifications}(b), typical chainsaw features are also made visible by the overlayed saliency map.

\subsection{Sanity check and robustness}
As a sanity check, following the method in \citet{adebayo2018sanity}, the PCAM saliency maps have been visualized at each step of a cascading randomization of a VGG16 network, from last to first layer. 
The purpose of this sanity check is to verify that the Poly-CAM methods do not work as edge detectors, and effectively relies on the actual weights of the model to derive class-specific saliency maps. All PCAM variants successfully passed the test, as shown in Appendix \ref{app:sanity_check}.
\textcolor{\modcolor}{
To evaluate the robustness of our explanation method, a sensitivity analysis has been run, following the methodology introduced in \citep{ghorbani2019interpretation} and \citep{yeh2019fidelity}. Results are presented in Appendix \ref{app:robustness}. They reveal that PCAM has a small explanation sensitivity
, similar to the ones obtained by other CAM-based methods, and one or two orders of magnitude below the sensitivities obtained by gradient-based and perturbation methods. 
}

\section{Conclusion}

This paper has introduced the Poly-CAM method, to produce high resolution saliency maps without relying on gradient backpropagation. Three variants of our Poly-CAM framework are investigated, depending on whether the values weighting the activation maps are obtained by masking or unveiling image pixels, or both.
Our experiments reveal that the combined strategy, i.e. PCAM$^\pm$, provides the more versatile solution with state of the art performances in term of faithfulness insertion-deletion metrics and outperforming current available methods in term of precision of visualization.
Despite our work is a valuable step towards a more explainable AI, there is still plenty of room for improvement in this domain. One of the questions raised by this work is related to the way the importance of a pixel should be quantified. 
Indeed, the importance of a group of pixels appears to be different when this group is removed or when it is inserted (for example the importance of contextual information is more important when removing it than when inserting it), which can not be properly reflected by a single saliency map. 

\section*{Acknowledgment}
We want to thank the authors of Zoom-CAM for their kind help in using their method.

\section*{Funding Statement}
The Research Foundation for Industry and Agriculture, National Scientific Research Foundation (FRIA-FNRS http://www.fnrs.be/index.php) funded this research as a grant attributed to Alexandre Englebert, consisting in PhD financing.

Computational resources have been provided by the supercomputing facilities of the Université catholique de Louvain (CISM/UCL) and the Consortium des Équipements de Calcul Intensif en Fédération Wallonie Bruxelles (CÉCI) funded by the Fond de la Recherche Scientifique de Belgique (F.R.S.-FNRS) under convention 2.5020.11 and by the Walloon Region

\subsubsection*{Reproducibility Statement}
Attention was ported on the reproductibility of this paper.
The source code for Poly-CAM is provided as supplementary material.
The list of images used from the 2012 ILSVRC validation set \citep{ILSVRC15} is also provided.
The saliency maps used in this paper can be generated as npz files using provided scripts, or alternatively can be downloaded from \url{https://polycam.ddns.net}, an anonymous website made for the blind review process. One npz file is provided for each saliency method applied to one model. The files are named based on the model used ("vgg16" or "resnet50") and the saliency method used. Inside each npz file, the saliency maps can be retrieve in a numpy array format using the name of the image (with the ".JPEG" extension included) as a key.
Similarly, measurement for the faithfulness metrics can either be generated as csv files or downloaded from \url{https://polycam.ddns.net}. The csv files are named as so '\{del\_,ins\_\}\_\{auc,details\}\_\{vgg16,resnet50\}\_[saliency].csv'. The prefix "del\_" and "ins\_" refer to files associated with the deletion and insertion metrics respectively, "auc" files contains the area under the curve for the above metrics while files named with "details" contains the class certitude for each step of insertion or deletion, "vgg16" and "resnet50" refers to the model and "saliency" is the name of the saliency method.
Jupyter notebooks are also provided for easier handling. Practical information can be found in the README.md file inside the supplementary zip file.

\bibliographystyle{plainnat}
\bibliography{references}

\newpage

\appendix

\section{Poly-CAM vs previous works}\label{app:methods_comparison_full}

\begin{figure}[H]
	\centering
	\includegraphics[width=\textwidth]{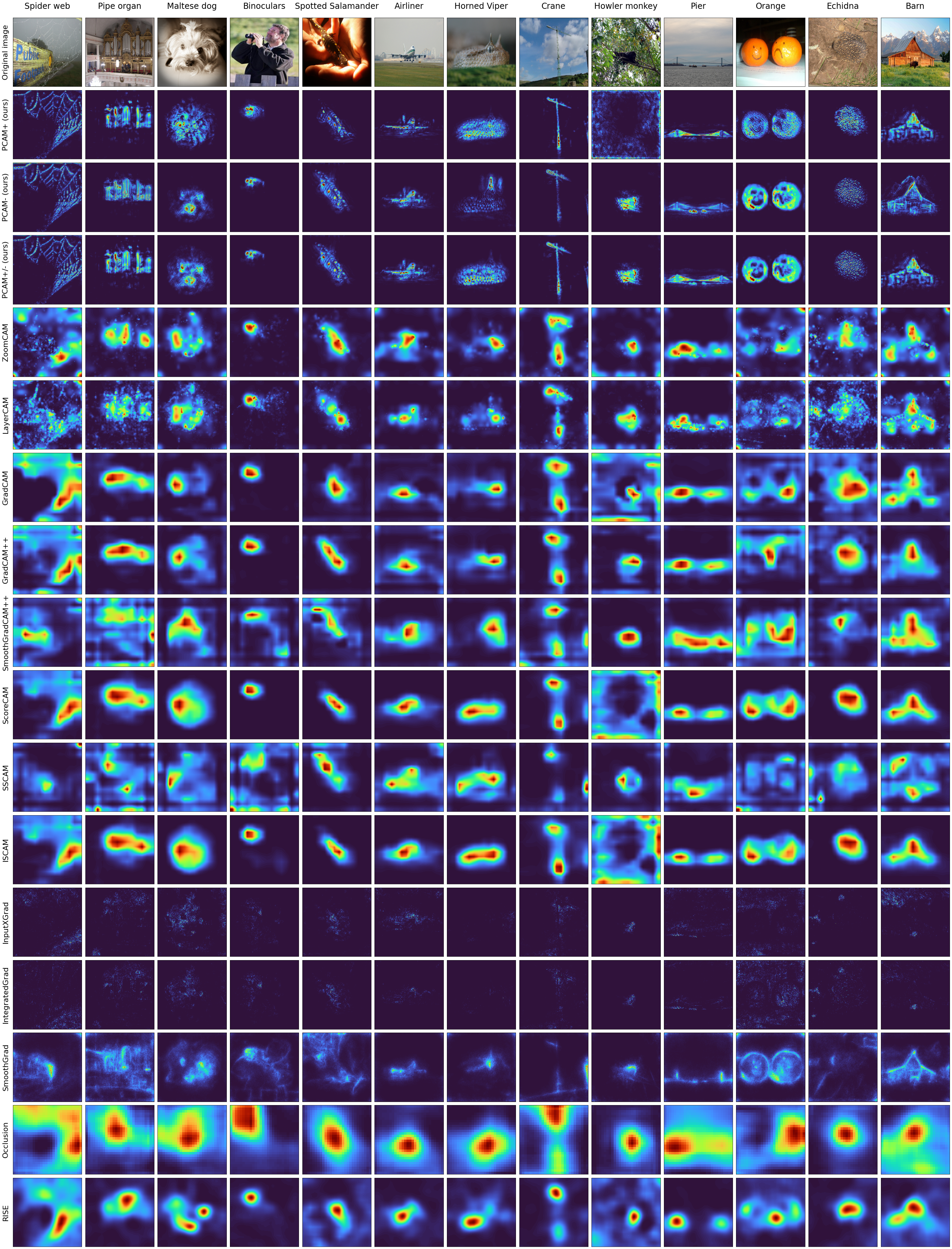}
	\caption{Visual comparison of methods. The compared methods are the three Poly-CAM variants proposed in this paper (PCAM$^+$, PCAM$^-$, PCAM$^\pm$), Zoom-CAM \citep{shi2021zoom}, Layer-CAM \citep{jiang2021layercam}, Grad-CAM \citep{selvaraju2017grad}, Grad-CAM++ \citep{chattopadhay2018grad}, Smooth Grad-CAM++ \citep{omeiza2019smooth}, Score-CAM \citep{wang2020score}, SS-CAM \citep{wang2020ss}, IS-CAM \citep{naidu2020cam}, Input X Gradient \citep{shrikumar2016not}, IntegratedGradient \citep{sundararajan2017axiomatic}, SmoothGrad \citep{smilkov2017smoothgrad}, Occlusion \citep{zeiler2014visualizing}, RISE \citep{petsiuk2018rise}.}
	\label{fig:methods_comparison_full}
\end{figure}

\section{\textcolor{\modcolor}{Ablation Study}}\label{app:ablation_study}
\color{\modcolor}

Section \ref{app:pcam_isolated_cam} presents an ablation study designed to assess the importance of combining different layers (as proposed in \eqref{eq:poly-cam}) versus deriving the saliency map from a single layer (as defined by \eqref{eq:cam}).
vs using a specific layer in isolation in section \ref{app:pcam_isolated_cam}. Section \ref{app:lnorm_ablation} reveals the critical importance of the LNorm operation in \eqref{eq:lnorms}.

\subsection{Poly-CAM versus Layer-specific map}\label{app:pcam_isolated_cam}

The interest of combining the activation maps from multiple layers, as proposed by \eqref{eq:poly-cam}, is demonstrated by comparing Poly-CAM with the saliency maps derived in individual layers, using \eqref{eq:cam}. The three weighting factors presented in Section \ref{weights_definition} are considered.

\begin{figure}[H]
	\centering
    \includegraphics[width=\linewidth]{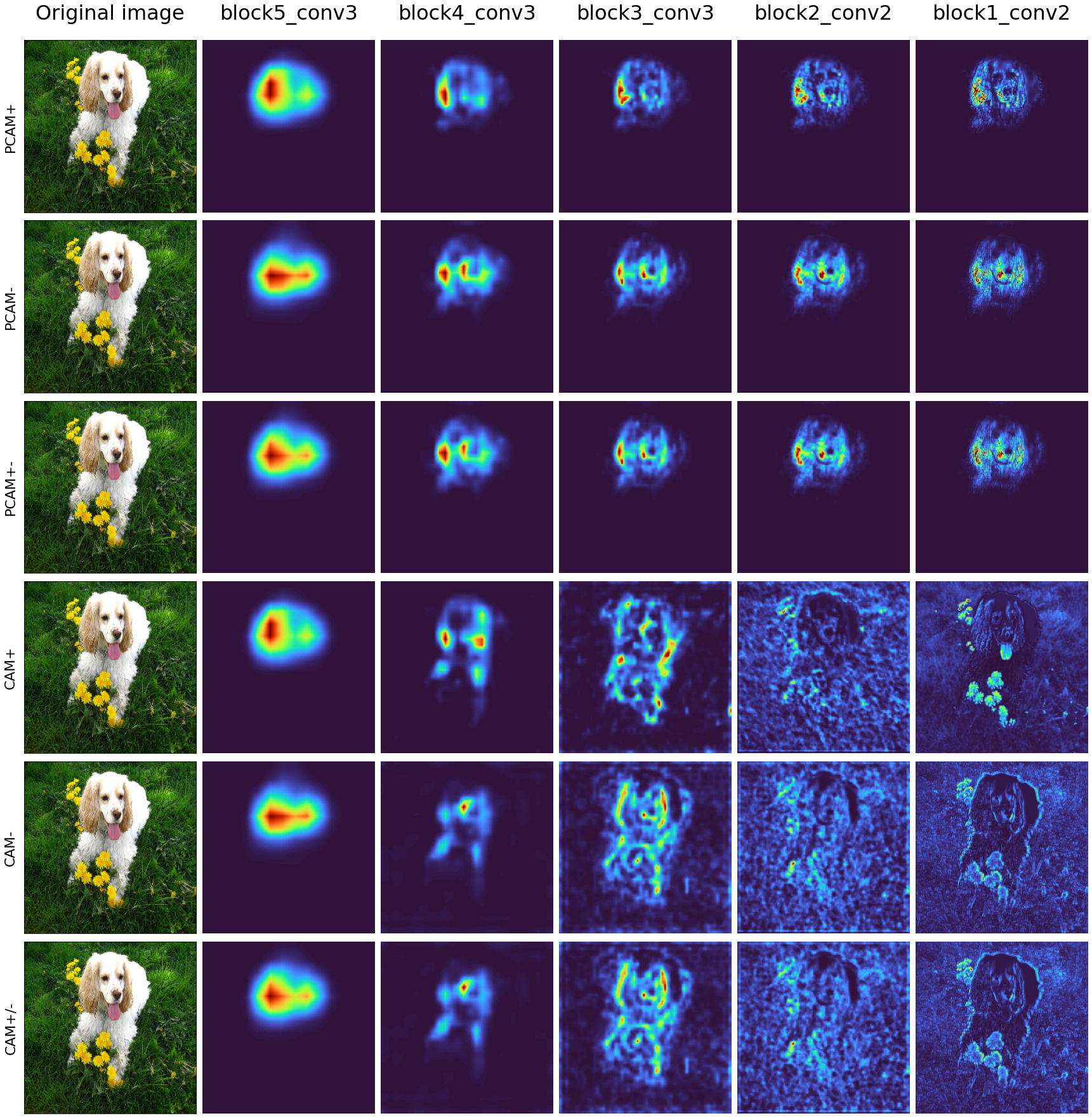}
    \caption{Comparison of Poly-CAM vs classical CAM-based results as a function of the layer, with the three proposed weights in Section \ref{weights_definition} on VGG16. We can observe that computing the CAM saliency map from a single and early layer gives poor results, with maps highlighting elements all over the input image. In contrast, Poly-CAM methods increase the resolution while keeping the focus of saliency on class-specific relevant features.}
    \label{fig:isolated_vs_pcam_cocker}
\end{figure}
\begin{figure}[H]
	\centering
    \includegraphics[width=\linewidth]{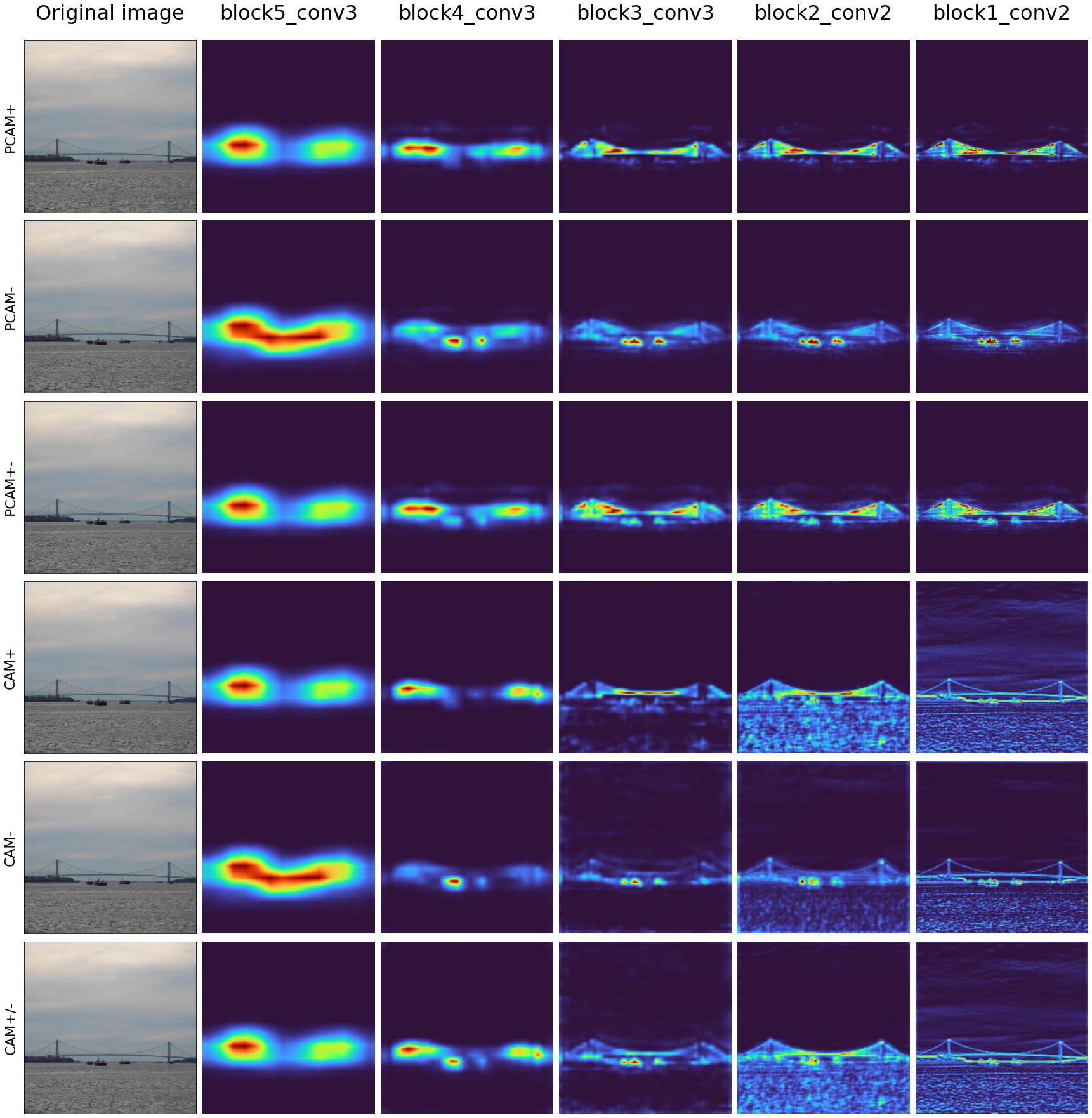}
    \caption{Comparison of Poly-CAM vs classical CAM-based results as a function of the layer, with the three proposed weights in Section \ref{weights_definition} on VGG16. We can observe that computing the CAM saliency map from a single and early layer gives poor results, with maps highlighting elements all over the input image. In contrast, Poly-CAM methods increase the resolution while keeping the focus of saliency on class-specific relevant features.}
    \label{fig:isolated_vs_pcam_pier}
\end{figure}
\begin{figure}[H]
	\centering
	\includegraphics[width=\textwidth]{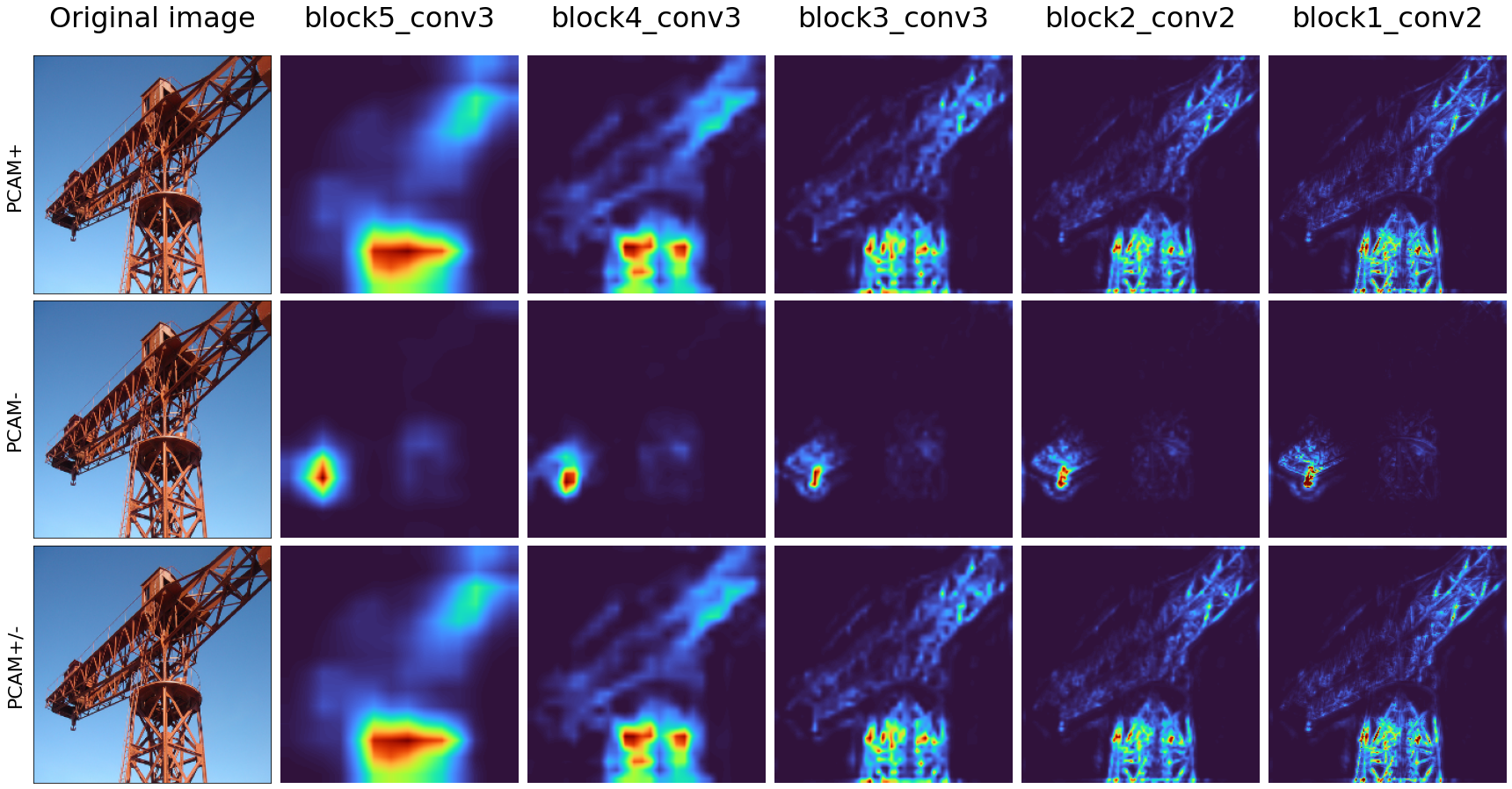}
	\caption{Comparison of Poly-CAM for different layers of VGG16 on a crane. The saliency map is refined as layers are added from left to right, showing more precise structures.
	We can also see that PCAM$^-$ gives more importance to the hook while PCAM$^+$ and PCAM$^\pm$ give more importance on the mest and jib of the crane. This reflects the fact that, given the metallic structure present in the image, removing the hook most significantly impact the crane functionality of this metallic structure. In contrast, all metallic parts are worth to be added when starting from a baseline image.}
	\label{fig:pcam_layers_crane}
\end{figure}
\begin{figure}[H]
	\centering
    \includegraphics[width=\textwidth]{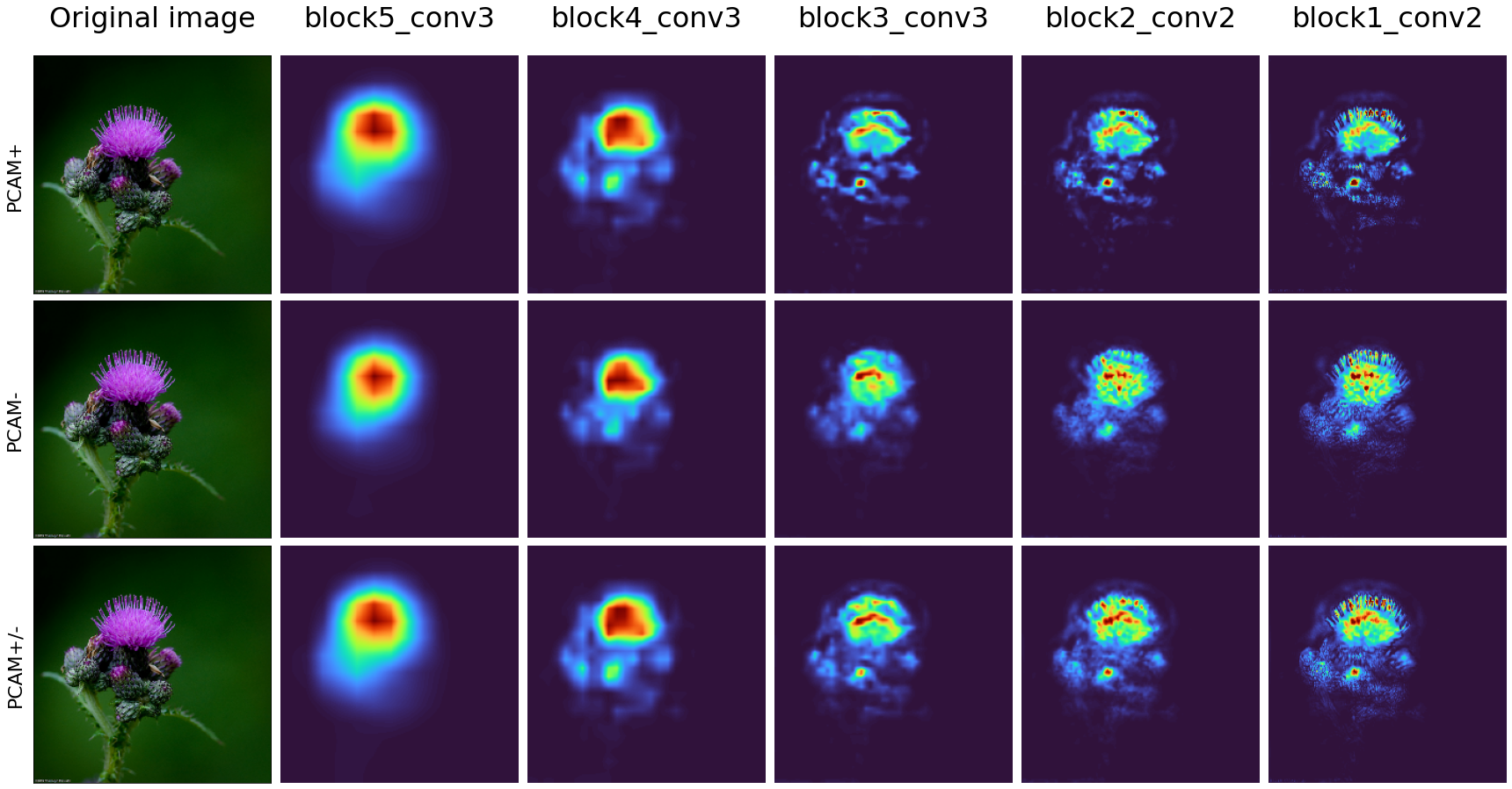}
    \caption{Comparison of Poly-CAM for different layers of VGG16 on a cardoon. The focus is progressively more intense in the spikes of the flowers, the most important part seems to be at the tips of the violet spikes for the three variants, which give similar results in this case.}
	\label{fig:pcam_layers_cardoon}
\end{figure}
\begin{figure}[H]
	\centering
    \includegraphics[width=\textwidth]{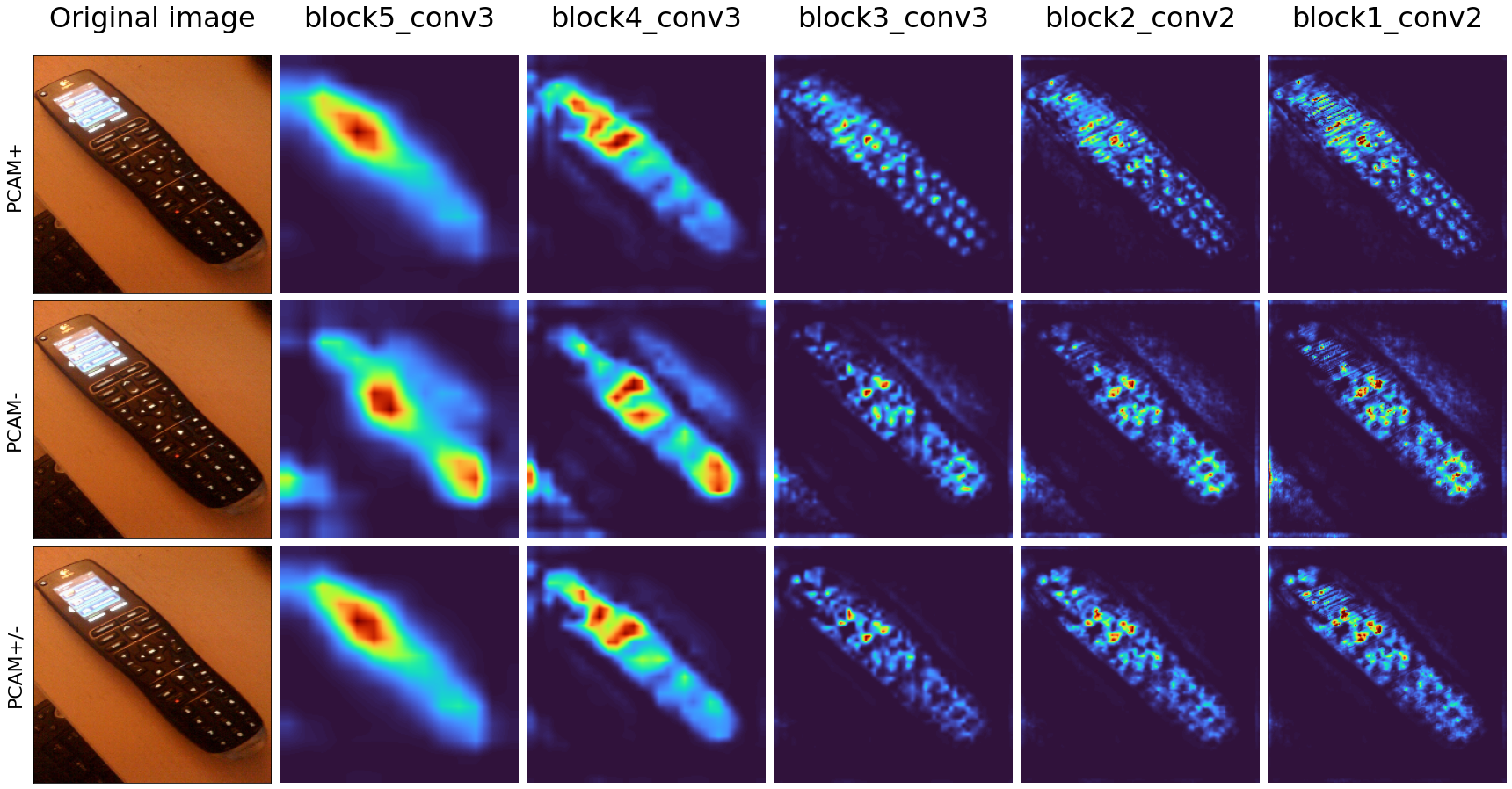}
    \caption{Comparison of Poly-CAM for different layers of VGG16 on a remote control.
    We can see the progressive focus on the buttons when we progress over the layers to produce a more precise map. In this case, the three variants produce similar results.}
	\label{fig:pcam_layers_remote_control}
\end{figure}
\begin{figure}[H]
	\centering
	\includegraphics[width=\textwidth]{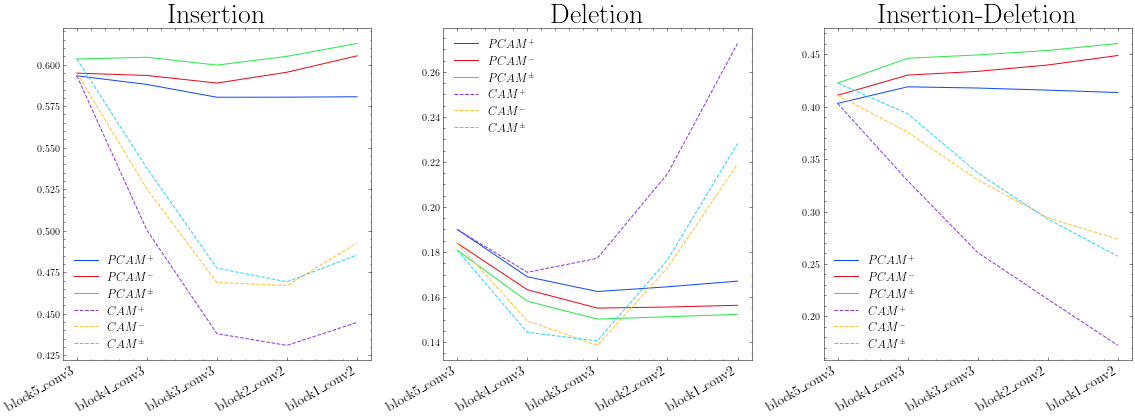}
	\caption{\color{\modcolor}Comparison of metrics as a function of the layer index for Poly-CAM vs CAM on isolated layers - VGG16.
	Last to first layer is represented left to right. This show that using CAM saliency maps on early layers performs poorly.}
	\label{graph:isolated_vs_pcam_vgg}
\end{figure}
\begin{figure}[H]
	\centering
	\includegraphics[width=\textwidth]{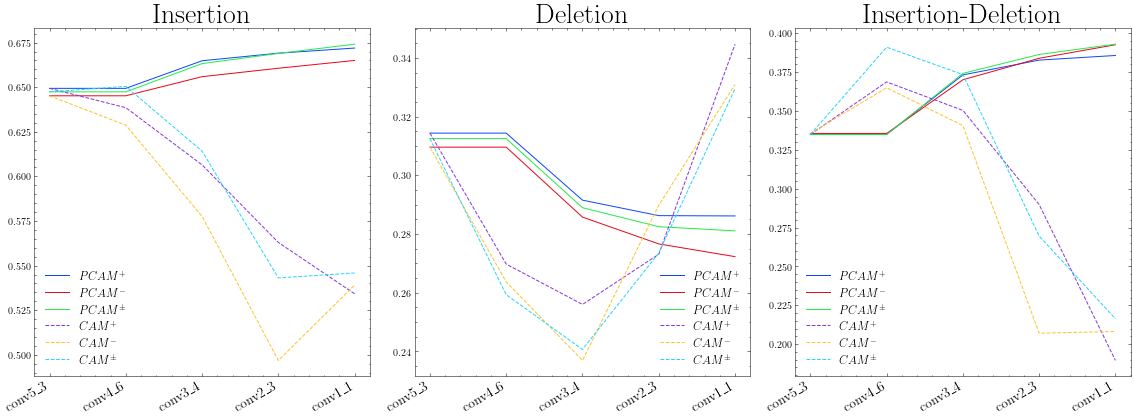}
	\caption{\color{\modcolor}Comparison of metrics as a function of the layer index for Poly-CAM vs CAM on isolated layers - ResNet50.
	Last to first layer is represented left to right. This show that using CAM saliency maps on early layers performs poorly.}
	\label{graph:isolated_vs_pcam_resnet}
\end{figure}

The set-up and models are the same as the ones used in Section \ref{dataset_methods} and the target layers are [block1\_conv2,\ block2\_conv2,\ block3\_conv3,\ block4\_conv3,\ block5\_conv3] for VGG16, and [conv1\_1,\ conv2\_3,\ conv3\_4,\ conv4\_6,\ conv5\_3] for ResNet50. A visual comparison of the PCAM methods vs intermediate CAM is shown in Figure \ref{fig:isolated_vs_pcam_pier} and Figure \ref{fig:isolated_vs_pcam_cocker}. A visual comparison of the three Poly-CAM variants for differents layers is shown in Figure \ref{fig:pcam_layers_crane}, Figure \ref{fig:pcam_layers_cardoon} and Figure \ref{fig:pcam_layers_remote_control}. Two comparatives graphs are provided in Figure \ref{graph:isolated_vs_pcam_vgg} and Figure \ref{graph:isolated_vs_pcam_resnet} for VGG16 and ResNet50 respectively. The results are shown in Table \ref{table:ablation_vgg} for VGG16, and in Table \ref{table:ablation_resnet} for ResNet50.

While the Poly-CAM methods improve when integrating more layers, the CAM dramatically looses in class-specificity when considering early layers in the networks. 
This highlight that combining the layers is a better approach than using CAM on isolated early layers using the proposed $w_{l,k}$ scores. This poor behavior of CAM saliency maps in early layers confirms the results provided by \citet{shi2021zoom} in their ablation study.

\newpage

\begin{table}[H]
\caption{Faithfulness metrics for Poly-CAM and CAM on isolated layers - VGG16}
\centering
\begin{tabular}{l|cccc}
\hline
\multirow{2}{*}{Method} & \multicolumn{4}{c}{VGG16}\\
\hhline{~----}
 & Layer &  Insertion     &   Deletion     &     Ins-Del \\ \hline
\hline
\multirow{5}{*}{$PCAM^+$}&block5\_conv3 & 0.59 & 0.19 & 0.40\\
 &block4\_conv3 & 0.59 & 0.17 & 0.42\\
 &block3\_conv3 & 0.58 & 0.16 & 0.42\\
 &block2\_conv2 & 0.58 & 0.16 & 0.42\\
 &block1\_conv2 & 0.58 & 0.17 & 0.41\\
\hline
\multirow{5}{*}{$PCAM^-$}&block5\_conv3 & 0.59 & 0.18 & 0.41\\
 &block4\_conv3 & 0.59 & 0.16 & 0.43\\
 &block3\_conv3 & 0.59 & 0.16 & 0.43\\
 &block2\_conv2 & 0.60 & 0.16 & 0.44\\
 &block1\_conv2 & 0.60 & 0.16 & 0.45\\
\hline
\multirow{5}{*}{$PCAM^{\pm}$}&block5\_conv3 & 0.60 & 0.18 & 0.42\\
 &block4\_conv3 & 0.60 & 0.16 & 0.45\\
 &block3\_conv3 & 0.60 & 0.15 & 0.45\\
 &block2\_conv2 & 0.60 & 0.15 & 0.45\\
 &block1\_conv2 & 0.61 & 0.15 & 0.46\\
\hline
\multirow{5}{*}{$CAM^+$}&block5\_conv3 & 0.59 & 0.19 & 0.40\\
 &block4\_conv3 & 0.50 & 0.17 & 0.33\\
 &block3\_conv3 & 0.44 & 0.18 & 0.26\\
 &block2\_conv2 & 0.43 & 0.21 & 0.22\\
 &block1\_conv2 & 0.44 & 0.27 & 0.17\\
\hline
\multirow{5}{*}{$CAM^-$}&block5\_conv3 & 0.59 & 0.18 & 0.41\\
 &block4\_conv3 & 0.53 & 0.15 & 0.38\\
 &block3\_conv3 & 0.47 & 0.14 & 0.33\\
 &block2\_conv2 & 0.47 & 0.17 & 0.29\\
 &block1\_conv2 & 0.49 & 0.22 & 0.27\\
\hline
\multirow{5}{*}{$CAM^{\pm}$}&block5\_conv3 & 0.60 & 0.18 & 0.42\\
 &block4\_conv3 & 0.54 & 0.14 & 0.39\\
 &block3\_conv3 & 0.48 & 0.14 & 0.34\\
 &block2\_conv2 & 0.47 & 0.18 & 0.29\\
 &block1\_conv2 & 0.49 & 0.23 & 0.26\\
\hline
\end{tabular}
\label{table:ablation_vgg}
\end{table}

\begin{table}[H]
\caption{Faithfulness metrics for Poly-CAM and CAM on isolated layers - ResNet50}
\centering
\begin{tabular}{l|cccc}
\hline
\multirow{2}{*}{Method} & \multicolumn{4}{c}{ResNet50}\\
\hhline{~----}
 & Layer &  Insertion     &   Deletion     &     Ins-Del \\ \hline
\hline
\multirow{5}{*}{$PCAM^+$}&conv5\_3 & 0.65 & 0.31 & 0.33\\
 &conv4\_6 & 0.65 & 0.31 & 0.33\\
 &conv3\_4 & 0.66 & 0.29 & 0.37\\
 &conv2\_3 & 0.67 & 0.29 & 0.38\\
 &conv1\_1 & 0.67 & 0.29 & 0.38\\
\hline
\multirow{5}{*}{$PCAM^-$}&conv5\_3 & 0.65 & 0.31 & 0.34\\
 &conv4\_6 & 0.65 & 0.31 & 0.34\\
 &conv3\_4 & 0.66 & 0.29 & 0.37\\
 &conv2\_3 & 0.66 & 0.28 & 0.38\\
 &conv1\_1 & 0.66 & 0.27 & 0.39\\
\hline
\multirow{5}{*}{$PCAM^{\pm}$}&conv5\_3 & 0.65 & 0.31 & 0.33\\
 &conv4\_6 & 0.65 & 0.31 & 0.33\\
 &conv3\_4 & 0.66 & 0.29 & 0.37\\
 &conv2\_3 & 0.67 & 0.28 & 0.39\\
 &conv1\_1 & 0.67 & 0.28 & 0.39\\
\hline
\multirow{5}{*}{$CAM^+$}&conv5\_3 & 0.65 & 0.31 & 0.33\\
 &conv4\_6 & 0.64 & 0.27 & 0.37\\
 &conv3\_4 & 0.61 & 0.26 & 0.35\\
 &conv2\_3 & 0.56 & 0.27 & 0.29\\
 &conv1\_1 & 0.53 & 0.34 & 0.19\\
\hline
\multirow{5}{*}{$CAM^-$}&conv5\_3 & 0.65 & 0.31 & 0.34\\
 &conv4\_6 & 0.63 & 0.26 & 0.36\\
 &conv3\_4 & 0.58 & 0.24 & 0.34\\
 &conv2\_3 & 0.50 & 0.29 & 0.21\\
 &conv1\_1 & 0.54 & 0.33 & 0.21\\
\hline
\multirow{5}{*}{$CAM^{\pm}$}&conv5\_3 & 0.65 & 0.31 & 0.33\\
 &conv4\_6 & 0.65 & 0.26 & 0.39\\
 &conv3\_4 & 0.61 & 0.24 & 0.37\\
 &conv2\_3 & 0.54 & 0.27 & 0.27\\
 &conv1\_1 & 0.55 & 0.33 & 0.22\\
\hline
\end{tabular}
\label{table:ablation_resnet}
\end{table}

\subsection{Importance of LNorm}\label{app:lnorm_ablation}

The importance of including the LNorm operator in equation (3) is challenged this section. We produced saliency maps using both the complete method and a variant where LNorm has been ablated. Formally, the saliency map of the LNorm ablated method is

\begin{equation}
\color{\modcolor}
PwithoutLNorm^c_l = 
\left \{
    \renewcommand\arraystretch{2}
    \begin{array}{ll}
        ReLU(\sum_{k} w_{l,k}(c) \mA^k_l ) & \mbox{for } l = L \\
        ReLU\left(\sum_{k} w_{l, k}(c) \mA^{k}_{l}  \right) \odot \uparrow_{bi}\left(\mP^c_{l+1}, \frac{s_{l+1}}{s_l}\right) & \mbox{for } 1 \leq l \leq L-1 \\
    \end{array}
\right .
\label{eq:poly-cam-nolnorm}
\end{equation}

\begin{figure}[H]
	\centering
	\begin{subfigure}{0.8\textwidth}
        \centering
        \includegraphics[width=\linewidth]{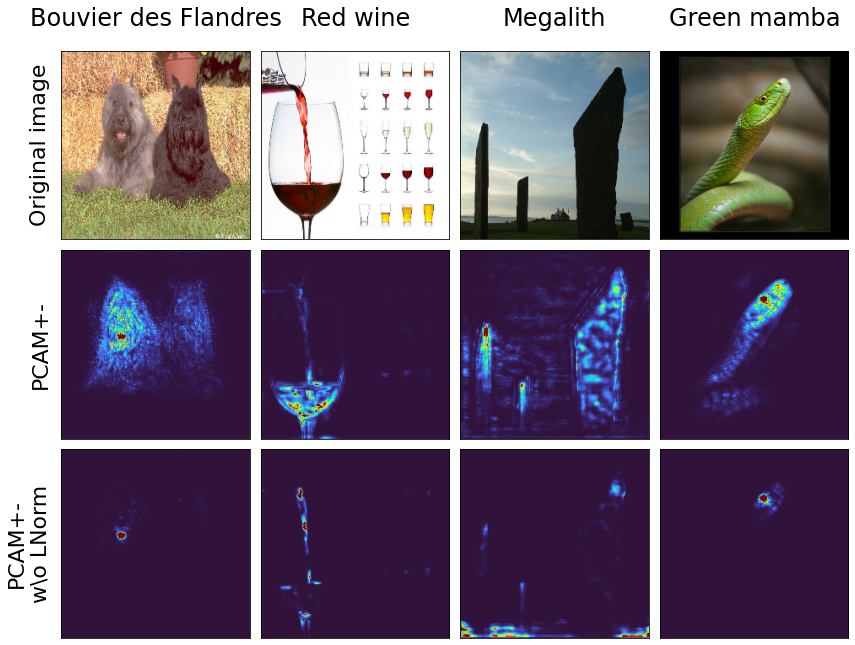}
        \vspace{0.2cm}
    \end{subfigure}
    \begin{subfigure}{0.8\textwidth}
        \centering
        \includegraphics[width=\linewidth]{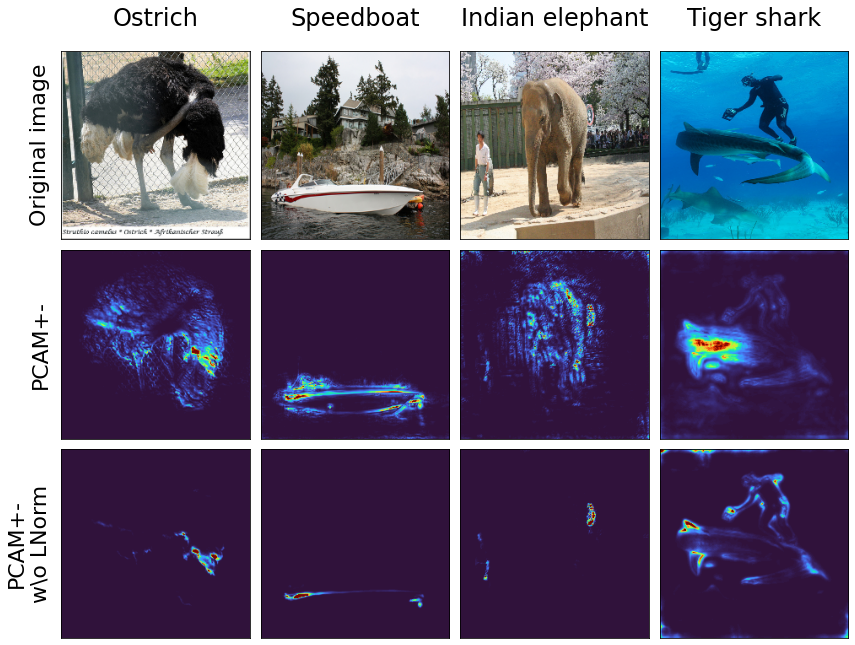}
    \end{subfigure}
	\caption{Visual comparison of PCAM$^{\pm}$ with and without LNorm. Without LNorm the visualisation tends to concentrate on very focal elements of the images like eyes, mouth,... Sometime some elements of the image that are not in object of the target class become also highlighted, like an object behind the elephant, or the diver next to the shark.}
	\label{fig:lnorm_visualisation}
\end{figure}

Representative examples are presented in Figure \ref{fig:lnorm_visualisation} to compare the conventional and ablated PCAM, when considering their high resolution saliency maps. We clearly observe that the ablated method tend to ignore some of class-relevant features, and focuses on a limited set of highly contrasted features (such as eye, mouth, or beak).



\color{black}

\section{Multiple classes comparison}\label{app:class_spe}

\begin{figure}[H]
	\centering
	\includegraphics[width=\textwidth]{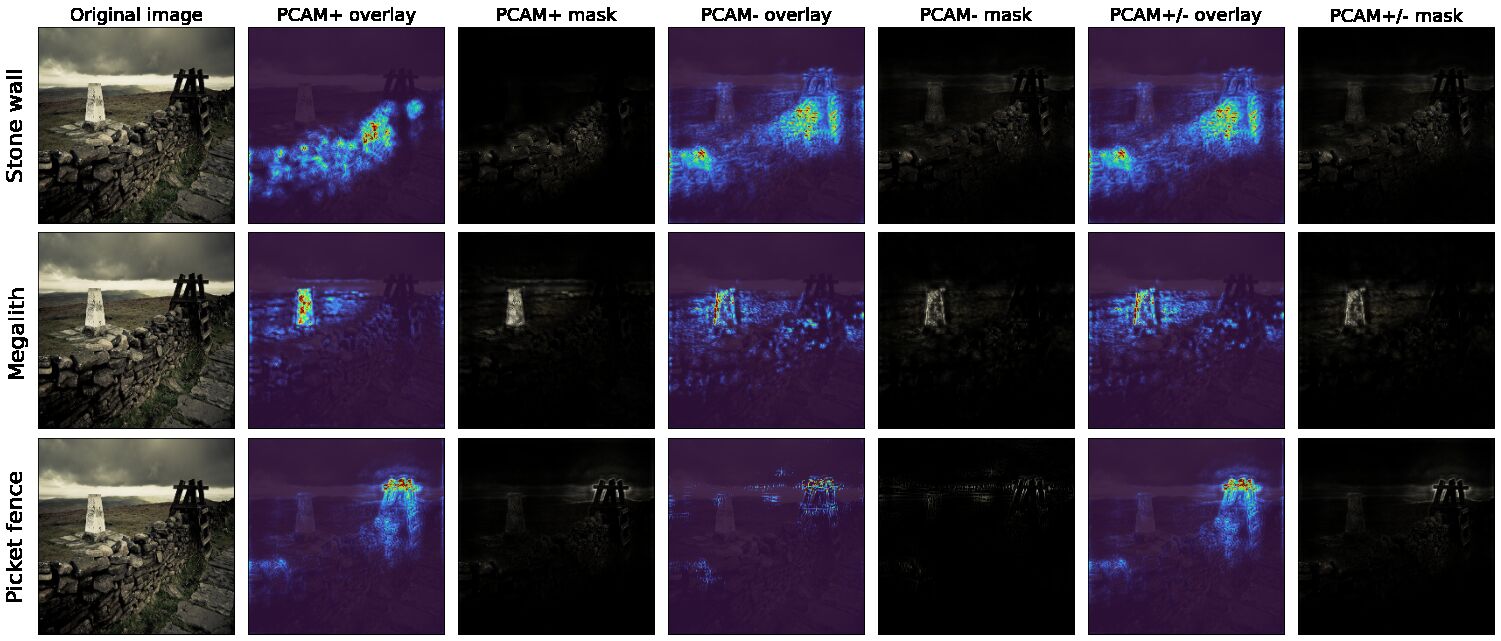}
	\caption{Comparison of multiple classes on an image with a stone wall, a megalith and pickets for the three Poly-CAM variants. The three classes are similarly separated for the three methods, while PCAM$^+$ being the more specific}
\end{figure}

\begin{figure}[H]
	\centering
	\includegraphics[width=\textwidth]{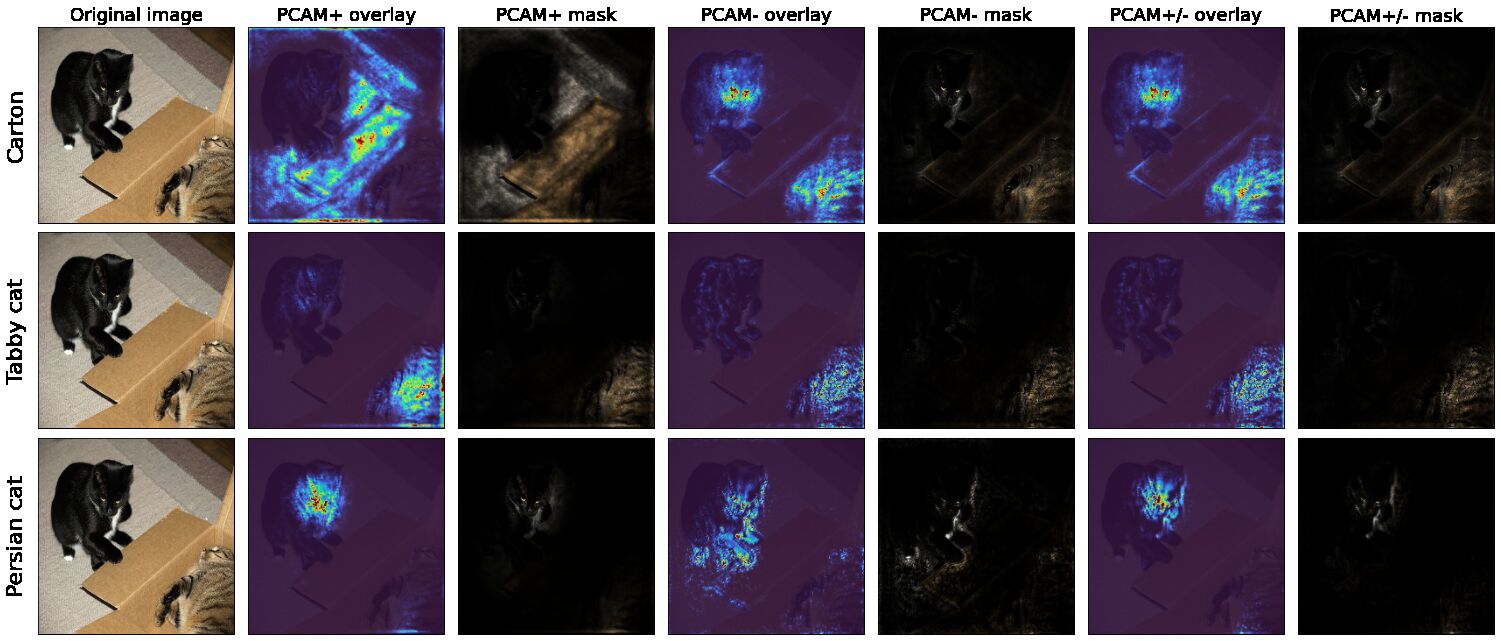}
	\caption{Comparison of multiple classes on an image with two different cats and a carton for the three Poly-CAM variants. The three methods correctly identify the two cats as different when using the Tabby cat and Persian cat classes, PCAM$^+$ correctly separate the carton while PCAM$^-$ and PCAM$^\pm$ fail for this class.}
\end{figure}

\section{Examples on Bone X-Ray}\label{app:xray}

\subsection{Cast bias visualization}\label{app:cast_bias}

\begin{figure}[H]
	\centering
	\begin{subfigure}{.4\textwidth}
        \centering
        \includegraphics[width=\linewidth]{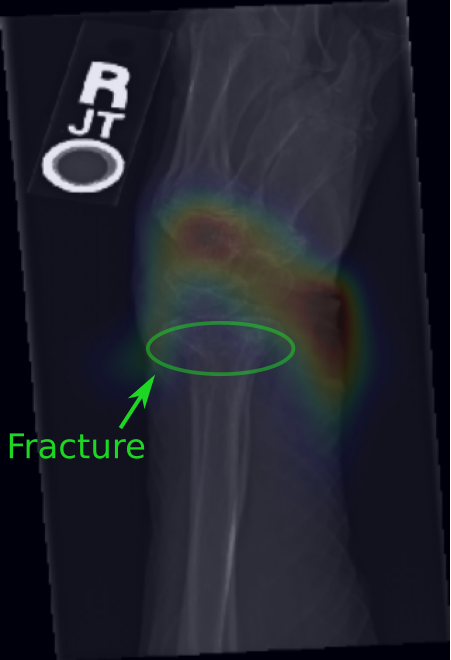}
        \label{fig:sub_xray1}
    \end{subfigure}
    \begin{subfigure}{.55\textwidth}
        \centering
        \includegraphics[width=\linewidth]{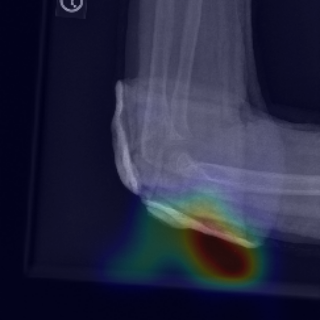}
        \label{fig:sub_xray2}
    \end{subfigure}
	\caption{Visualization of a Class Activation Map \citep{zhou2016learning} for the pathological label on two bone X-Ray images from MURA dataset \citep{rajpurkar2017mura}. Both images are labeled as pathological by the model, but only the left image show a bone fracture (manually annotated in green), the right image is clean of any fracture. The model seems to rely mostly on the plaster rather than on the absence of fracture to make a decisions. The model is a ResNet50 \citep{he2016deep} initialised on ImageNet, trained on the MURA dataset \citep{rajpurkar2017mura} for 50 epochs with Adam optimizer, an initial learning rate of 6e-5 with a cosine Annealing scheduler without restart, weight decay at 1e-5. Images are resized to 320x320 with random rotation up to 15° during training.}
	\label{fig:plaster_xray}
\end{figure}

\subsection{Importance of being accurate when localizing saliency}\label{app:xray_acc}

\begin{figure}[H]
	\centering
    \includegraphics[width=\linewidth]{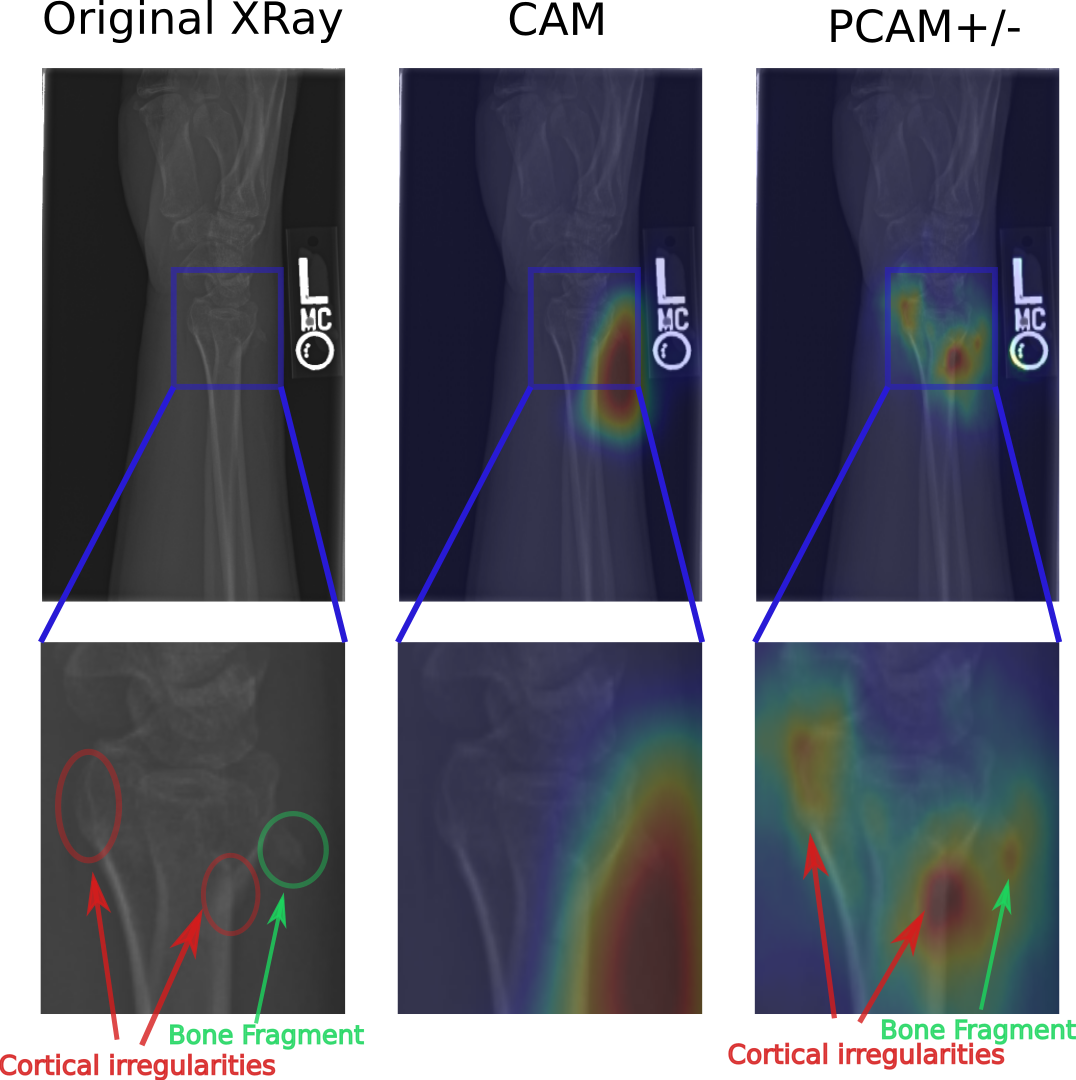}
    \caption{Visual comparison of Class Activation Map \citep{zhou2016learning} and PCAM$^\pm$ on a XRay of a bone fracture from MURA dataset \citep{rajpurkar2017mura}, for the pathological class label. The bottom row is a zoom on the fracture area. Manual annotations for cortical irregularities and bone fragments (the main signs of the presence of a fracture on this XRay) are shown in red and green ovals. The Class Activation Map is not precise, it seems to include the bone fragment and the right cortical irregularity but due to the low resolution, the highlighted area is very large and go far from the fracture. In comparison, PCAM$^\pm$ highlight smaller structures and seems to identify correctly the cortical irregularities and the bone fragment on this image, being probably a greater help for a physician.
    The model is a ResNet50 \citep{he2016deep} initialised on ImageNet, trained on the MURA dataset \citep{rajpurkar2017mura} for 50 epochs with Adam optimizer, an initial learning rate of 6e-5 with a cosine Annealing scheduler without restart, weight decay at 1e-5. Images are resized to 320x320 with random rotation up to 15° during training.}
	\label{fig:xray_acc}
\end{figure}

\section{Faithfulness metrics: supplementary data}

\subsection{Table}

\begin{table}[H]
\color{\modcolor}
\caption{Faithfulness metrics for all methods: CAM-based, gradient and perturbation methods}
\centering
\begin{tabular}{l|ccc|ccc}
\hline
\multirow{2}{*}{Methods} & \multicolumn{3}{c}{VGG16} & \multicolumn{3}{c}{ResNet50}    \\ \hhline{~------}
                &  Insertion     &   Deletion     &     Ins-Del    &  Insertion     &   Deletion    & Ins-Del    \\ \hline
                
IntegratedGradient & 0.41 & \textbf{0.10} & 0.31 & 0.52 & \textbf{0.16} & 0.36\\
InputXGrad & 0.37 & 0.12 & 0.26 & 0.47 & 0.18 & 0.28\\
SmoothGrad & 0.54 & 0.20 & 0.34 & 0.62 & 0.29 & 0.33\\\hline
RISE & \textbf{0.62} & 0.18 & 0.44 & \textbf{0.67} & 0.28 & \textbf{0.39}\\
Occlusion & \textbf{0.62} & 0.23 & 0.39 & 0.66 & 0.33 & 0.33\\\hline
GradCAM & 0.58 & 0.18 & 0.40 & 0.65 & 0.31 & 0.35\\
GradCAM++ & 0.57 & 0.19 & 0.38 & 0.65 & 0.31 & 0.34\\
SmoothGradCAM++ & 0.54 & 0.21 & 0.33 & 0.63 & 0.32 & 0.30\\
ScoreCAM & 0.59 & 0.19 & 0.40 & 0.65 & 0.31 & 0.34\\
SSCAM & 0.50 & 0.23 & 0.27 & 0.59 & 0.36 & 0.24\\
ISCAM & 0.59 & 0.19 & 0.40 & 0.65 & 0.32 & 0.33\\
ZoomCAM & 0.60 & 0.14 & \textbf{0.46} & 0.66 & 0.29 & 0.37\\
LayerCAM & 0.58 & 0.14 & 0.44 & 0.65 & 0.30 & 0.35\\\hline
PCAM$^+$ (ours) & 0.58 & 0.17 & 0.41 & \textbf{0.67} & 0.29 & 0.38\\
PCAM$^-$ (ours) & 0.60 & 0.16 & 0.45 & 0.66 & \textbf{0.27} & \textbf{0.39}\\
PCAM$^{\pm}$ (ours) & \textbf{0.61} & 0.15 & \textbf{0.46} & \textbf{0.67} & 0.28 & \textbf{0.39}\\
\hline
\end{tabular}
\label{table_comparison_other}
\caption*{{Insertion (higher is better), deletion (lower is better) and Insertion-Deletion (higher is better) with VGG16 and ResNet50 on the 2012 ILSVRC validation set. Comparison of our Poly-CAM methods with gradient methods: Input X Gradient \citep{shrikumar2016not}, Integrated Gradient \citep{sundararajan2017axiomatic}, SmoothGrad \citep{smilkov2017smoothgrad}, perturbation methods: Occlusion \citep{zeiler2014visualizing} and RISE \citep{petsiuk2018rise}, and CAM methods: Grad-CAM \citep{selvaraju2017grad}, Grad-CAM++ \citep{chattopadhay2018grad}, Score-CAM \citep{wang2020score}, SS-CAM \citep{wang2020ss}, IS-CAM \citep{naidu2020cam}, Smooth Grad-CAM++ \citep{omeiza2019smooth}, Zoom-CAM \cite{shi2021zoom} and Layer-CAM \citep{jiang2021layercam}. The grad methods are generally better in deletion but worst in insertion, translation by noisy saliency maps. Perturbations methods are similar to PCAM$^\pm$ in insertion, with a better result in VGG16 and lower result in ResNet50, but are less effective in deletion for both models. PCAM$^\pm$ give better average results, similar to Zoom-CAM for VGG16. For ResNet50 the results are better over all methods.}}
\end{table}

\subsection{Curves}\label{app:curves}

\begin{figure}[H]
	\centering
	\includegraphics[width=\textwidth]{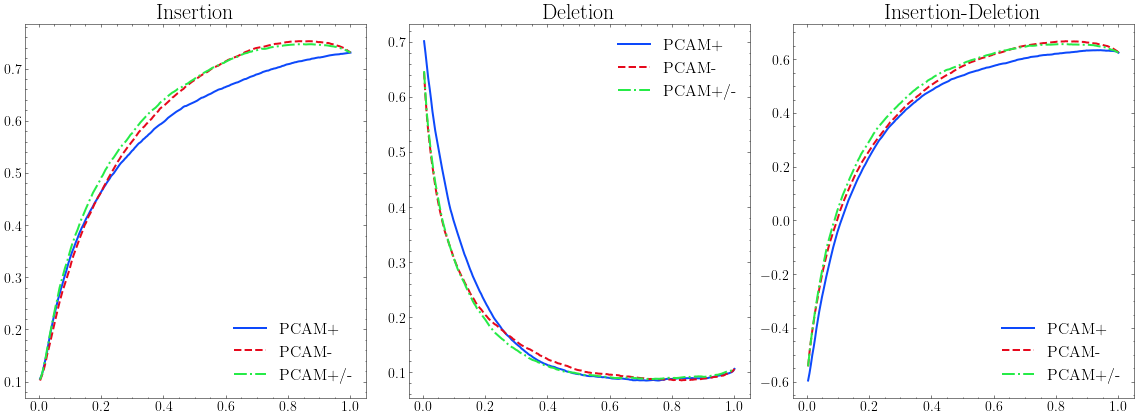}
	\caption{Faithfulness curves for Poly-CAM methods with VGG16 on the 2012 ILSVRC validation set. Comparison of the three variants of Poly-CAM.}
\end{figure}

\begin{figure}[H]
	\centering
	\includegraphics[width=\textwidth]{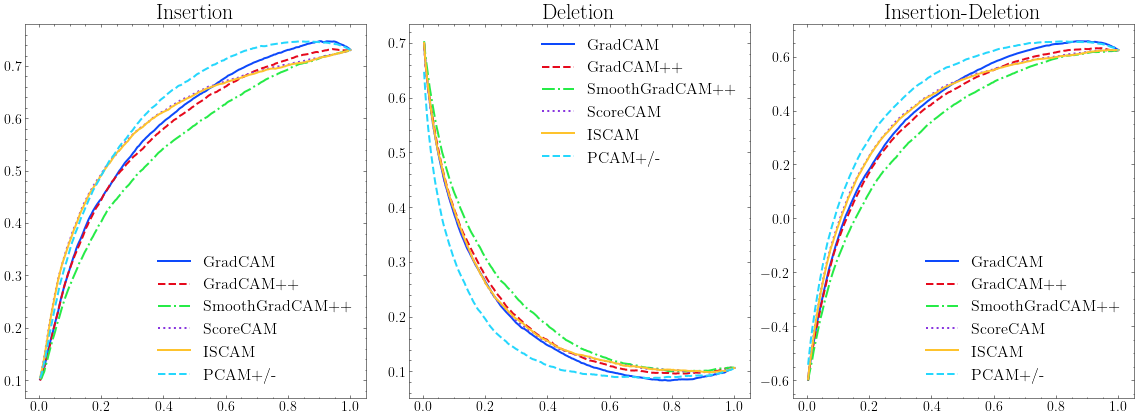}
	\caption{Faithfulness curves for CAM-based methods with VGG16 on the 2012 ILSVRC validation set. Comparison of our Poly-CAM methods with Grad-CAM \citep{selvaraju2017grad}, Grad-CAM++ \citep{chattopadhay2018grad}, Score-CAM \citep{wang2020score}, IS-CAM \citep{naidu2020cam}, Smooth Grad-CAM++ \citep{omeiza2019smooth}. SS-CAM \citep{wang2020ss} was excluded in this graph for ease of view}
	\label{graph:vgg_cam}
\end{figure}

\begin{figure}
	\centering
	\includegraphics[width=\textwidth]{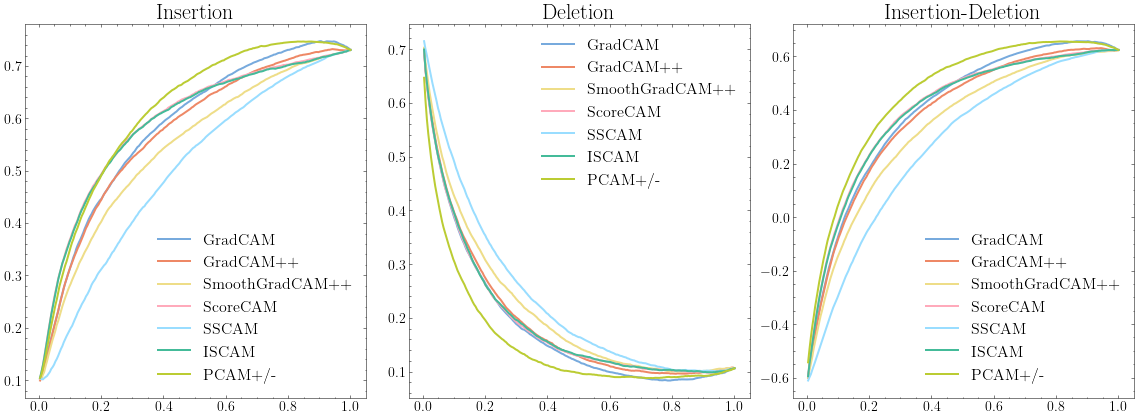}
	\caption{Faithfulness curves for CAM-based methods with VGG16 on the 2012 ILSVRC validation set. Comparison of our Poly-CAM methods with Grad-CAM \citep{selvaraju2017grad}, Grad-CAM++ \citep{chattopadhay2018grad}, Score-CAM \citep{wang2020score}, SS-CAM \citep{wang2020ss}, IS-CAM \citep{naidu2020cam}, Smooth Grad-CAM++ \citep{omeiza2019smooth}}
\end{figure}

\begin{figure}[H]
	\centering
	\includegraphics[width=\textwidth]{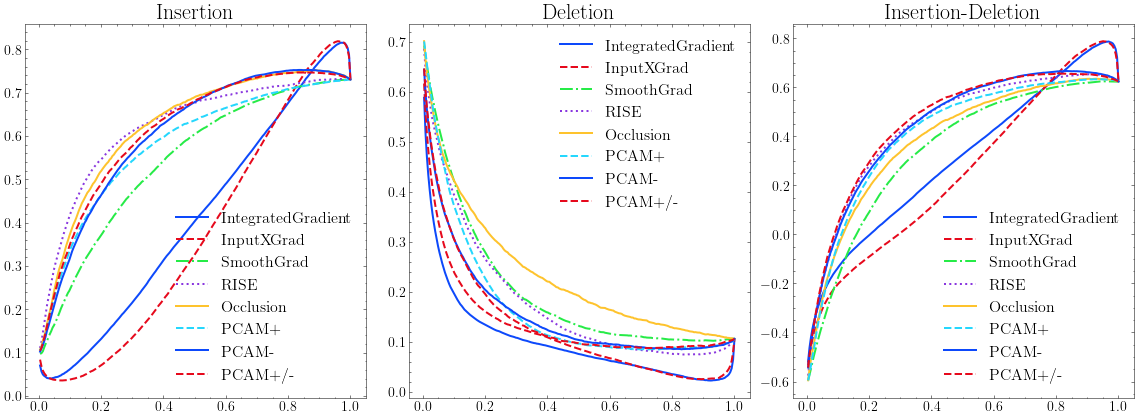}
	\caption{Faithfulness curves for gradient and perturbation methods with VGG16 on the 2012 ILSVRC validation set. Comparison of our PCAM$^\pm$ methods with gradient methods: Input X Gradient \citep{shrikumar2016not}, Integrated Gradient \citep{sundararajan2017axiomatic}, SmoothGrad \citep{smilkov2017smoothgrad}, and perturbation methods: Occlusion \citep{zeiler2014visualizing} and RISE \citep{petsiuk2018rise}}
	\label{graph:vgg_others}
\end{figure}

\begin{figure}[H]
	\centering
	\includegraphics[width=\textwidth]{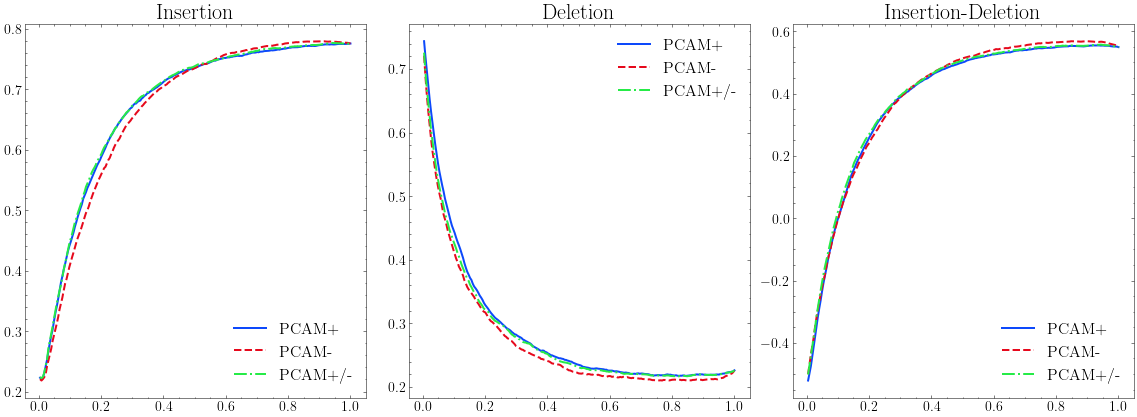}
	\caption{Faithfulness curves for Poly-CAM methods with ResNet50 on the 2012 ILSVRC validation set. Comparison of the three variants of Poly-CAM}
\end{figure}

\begin{figure}[H]
	\centering
	\includegraphics[width=\textwidth]{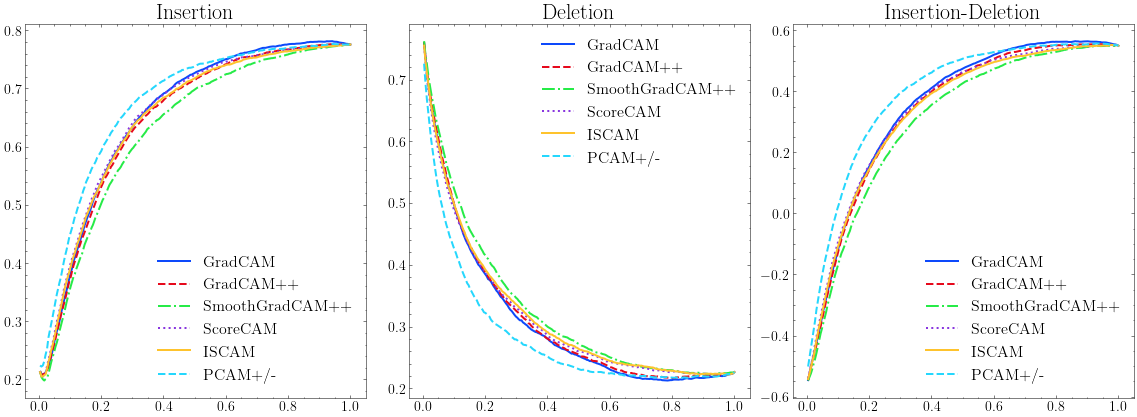}
	\caption{Faithfulness curves for CAM-based methods with ResNet50 on the 2012 ILSVRC validation set. Comparison of our Poly-CAM methods with Grad-CAM \citep{selvaraju2017grad}, Grad-CAM++ \citep{chattopadhay2018grad}, Score-CAM \citep{wang2020score}, IS-CAM \citep{naidu2020cam}, Smooth Grad-CAM++ \citep{omeiza2019smooth}. SS-CAM \citep{wang2020ss} was excluded in this graph for ease of view}
\end{figure}

\begin{figure}
	\centering
	\includegraphics[width=\textwidth]{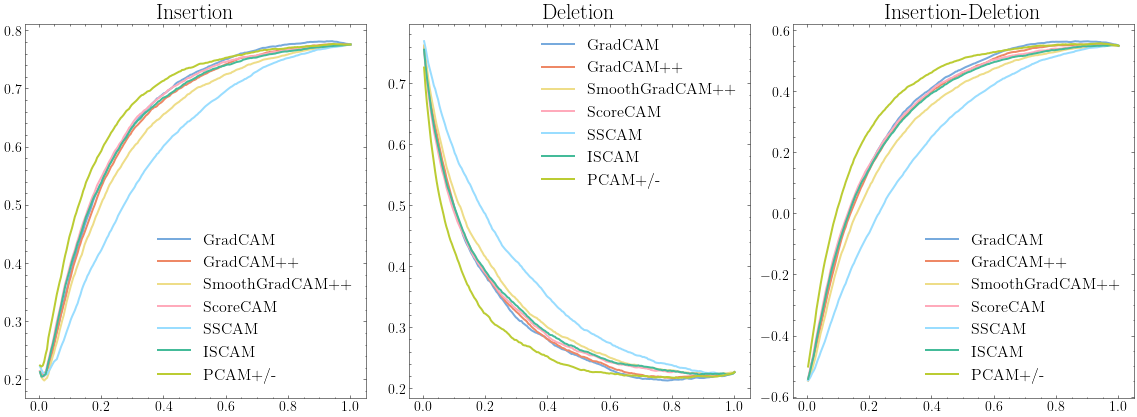}
	\caption{Faithfulness curves for CAM-based methods with ResNet50 on the 2012 ILSVRC validation set. Comparison of our Poly-CAM methods with Grad-CAM \citep{selvaraju2017grad}, Grad-CAM++ \citep{chattopadhay2018grad}, Score-CAM \citep{wang2020score}, SS-CAM \citep{wang2020ss}, IS-CAM \citep{naidu2020cam}, Smooth Grad-CAM++ \citep{omeiza2019smooth}}
\end{figure}

\begin{figure}[H]
	\centering
	\includegraphics[width=\textwidth]{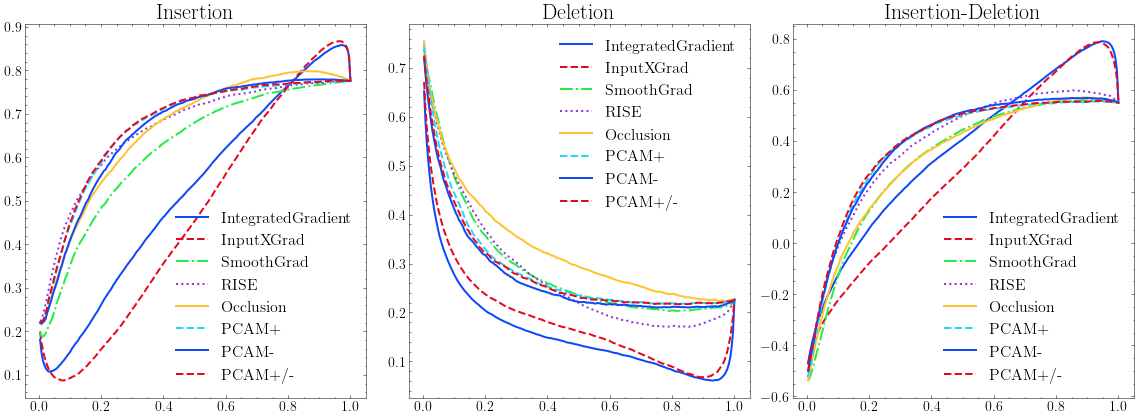}
	\caption{Faithfulness curves for gradient and perturbation methods with ResNet50 on the 2012 ILSVRC validation set. Comparison of our PCAM$^\pm$ methods with gradient methods: Input X Gradient \citep{shrikumar2016not}, Integrated Gradient \citep{sundararajan2017axiomatic}, SmoothGrad \citep{smilkov2017smoothgrad}, and perturbation methods: Occlusion \citep{zeiler2014visualizing} and RISE \citep{petsiuk2018rise}}
\end{figure}

\newpage

\section{Robustness}\label{app:robustness}
\color{\modcolor}

\begin{table}[H]
\caption{Sensitivity max table}
\centering
\begin{tabular}{l|c|c}
\hline
\multirow{2}{*}{Method} & \multicolumn{2}{c}{ Sensitivity max}\\
\hhline{~--}
 &VGG16&ResNet50\\ \hline
IntegratedGradient  & 0.3576 & 0.5299\\
InputXGrad & 0.6132 & 0.7225\\
SmoothGrad & 5.6824 & 7.7777\\ \hline
RISE & 0.7864 & 0.7841\\
Occlusion & 2.5176 & 3.4378\\ \hline
GradCAM & 0.0625 & 0.0212\\
GradCAM++ & 0.0525 & 0.0199\\
SmoothGradCAM++ & 0.5451 & 0.1594\\
ScoreCAM & 0.0466 & 0.0193\\
ISCAM & 0.0433 & 0.0334\\
ZoomCAM & 0.0987 & 0.0485\\
LayerCAM & 0.0937 & 0.0590\\ \hline
PCAM$^+$ (Ours) & 0.0650 & 0.0262\\
PCAM$^-$ (Ours) & 0.0837 & 0.0578\\
PCAM$^{\pm}$ (Ours) & 0.0659 & 0.0659\\
\hline
\end{tabular}
\label{table:sensitivity}
\caption*{\color{\modcolor}Sensitivity max metric measures maximum sensitivity of an explanation using Monte Carlo sampling-base approximation \citep{yeh2019fidelity}. Captum implementation was used with 10 perturbations per input and a epsilon radius of a L-Infinity ball set to 0.02 for sampling (defaults parameters from the implementation) \citep{kokhlikyan2020captum}. The compared methods are the three Poly-CAM variants proposed in this paper (PCAM$^+$, PCAM$^-$, PCAM$^\pm$), Zoom-CAM \citep{shi2021zoom}, Layer-CAM \citep{jiang2021layercam}, Grad-CAM \citep{selvaraju2017grad}, Grad-CAM++ \citep{chattopadhay2018grad}, Smooth Grad-CAM++ \citep{omeiza2019smooth}, Score-CAM \citep{wang2020score}, SS-CAM \citep{wang2020ss}, IS-CAM \citep{naidu2020cam}, Input X Gradient \citep{shrikumar2016not}, IntegratedGradient \citep{sundararajan2017axiomatic}, SmoothGrad \citep{smilkov2017smoothgrad}, Occlusion \citep{zeiler2014visualizing}, RISE \citep{petsiuk2018rise}.}
\end{table}

\color{black}

\newpage
\section{Sanity check}\label{app:sanity_check}

\begin{figure}[H]
	\centering
	\includegraphics[width=\textwidth]{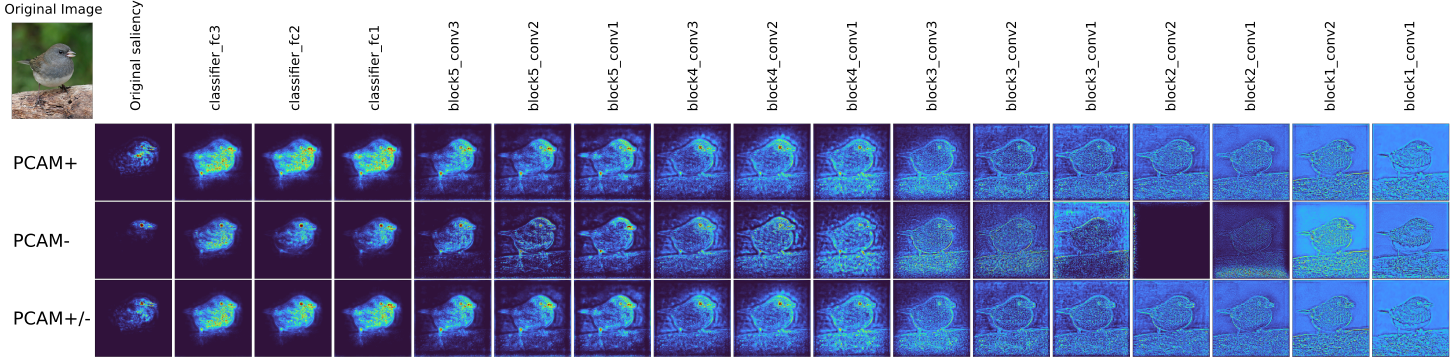}
	\caption{\textbf{Cascading randomization of VGG16.} Sanity check on Poly-CAM methods \citep{adebayo2018sanity}. Progression from left to right show a complete randomization of the VGG16 model, starting by the last layer up to the first layer. The methods are sensible to model randomization, which mean they pass this sanity check. It is interesting to note that the class specificity is lost rapidly after randomising the first classifier layer, then more and more features are lost while randomization progress up to the first layer of the network}
	\label{fig:sanity_cascading_bird}
\end{figure}

\begin{figure}[H]
	\centering
	\includegraphics[width=\textwidth]{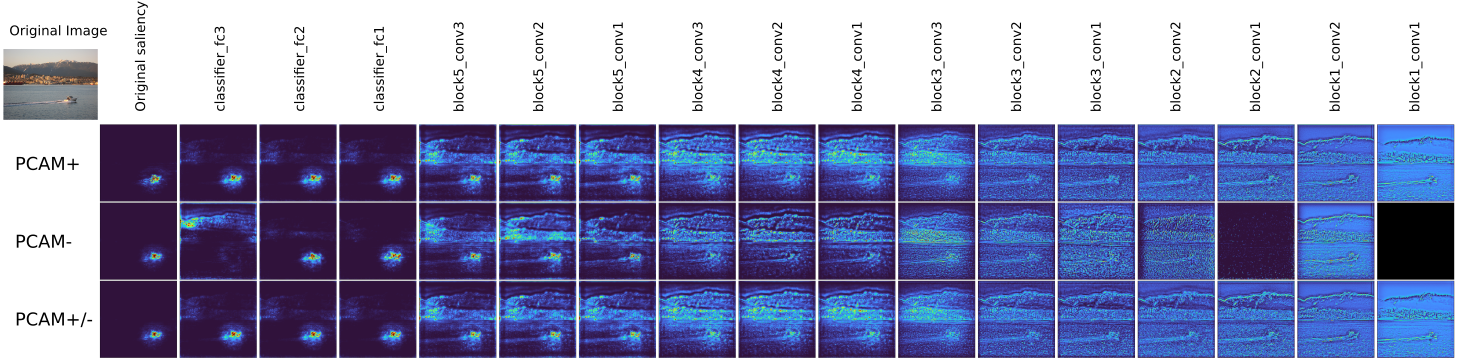}
	\caption{\textbf{Cascading randomization of VGG16.} Sanity check on Poly-CAM methods \citep{adebayo2018sanity}. Progression from left to right show a complete randomization of the VGG16 model, starting by the last layer up to the first layer.}
	\label{fig:sanity_cascading_boat}
\end{figure}

\begin{figure}[H]
	\centering
	\includegraphics[width=\textwidth]{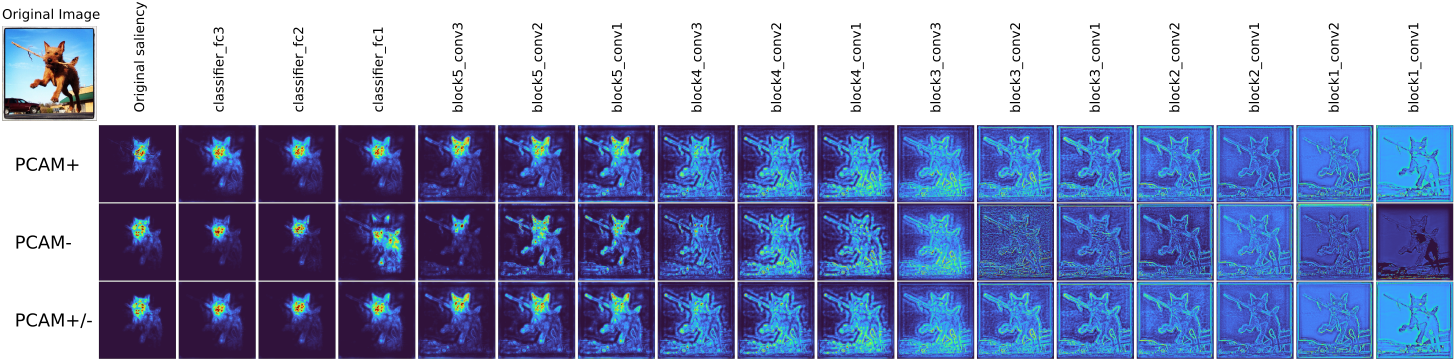}
	\caption{\textbf{Cascading randomization of VGG16.} Sanity check on Poly-CAM methods \citep{adebayo2018sanity}. Progression from left to right show a complete randomization of the VGG16 model, starting by the last layer up to the first layer.}
	\label{fig:sanity_cascading_dog}
\end{figure}

\end{document}